\documentclass[letterpaper, 10 pt, conference]{ieeeconf}

\usepackage{times}    
\usepackage{xspace}
\usepackage[dvipsnames, table]{xcolor}

\usepackage{amssymb}
\usepackage{booktabs}
\usepackage{graphicx}
\usepackage{diagbox}
\usepackage{multirow}
\usepackage{amsmath}
\usepackage{arydshln} 
\usepackage{tabularx}
\usepackage{algorithm}
\usepackage[noend]{algpseudocode}
\usepackage{pifont}
\usepackage{colortbl}
\usepackage{makecell}
\usepackage{comment}
\usepackage{pgfplots,tikz}
\pgfplotsset{compat=1.18}

\makeatletter
\let\NAT@parse\undefined
\makeatother

\usepackage[numbers,sort&compress]{natbib}

\definecolor{iccvblue}{rgb}{0.21,0.49,0.74}
\usepackage[breaklinks,colorlinks,allcolors=iccvblue]{hyperref}

\usepackage[capitalize]{cleveref}
\crefname{section}{Sec.}{Secs.}
\Crefname{section}{Section}{Sections}
\Crefname{table}{Table}{Tables}
\crefname{table}{Tab.}{Tabs.}

\colorlet{colorFst}{Green!25}       
\colorlet{colorSnd}{SpringGreen!45} 
\colorlet{colorTrd}{Yellow!30}      
\colorlet{colorLow}{darkgray!30}    
\newcommand{\fs}{\cellcolor{colorFst}\bf}   
\newcommand{\nd}{\cellcolor{colorSnd}}      
\newcommand{\rd}{\cellcolor{colorTrd}}      

\newcommand{\cmark}{{\color{PineGreen}\checkmark}}
\newcommand{\xmark}{{\color{red}\ding{55}}\ }

\DeclareTextFontCommand{\bemph}{\bfseries\em}
\newcommand{\boldparagraph}[1]{\vspace{1pt}\noindent{\bf #1}}

\newcommand{\ColorMapCircle}[1]{\textcolor{#1}{\ding{108}}}

\definecolor{r_wall}{RGB}{196, 51, 182}
\definecolor{r_ceiling}{RGB}{174, 199, 232}
\definecolor{r_floor}{RGB}{188, 189, 34}
\definecolor{r_chair}{RGB}{152, 223, 138}
\definecolor{r_blinds}{RGB}{255, 152, 150}
\definecolor{r_sofa}{RGB}{214, 39, 40}
\definecolor{r_table}{RGB}{91, 135, 229}
\definecolor{r_rug}{RGB}{31, 119, 180}
\definecolor{r_window}{RGB}{229, 91, 104}
\definecolor{r_lamp}{RGB}{247, 182, 210}
\definecolor{r_door}{RGB}{91, 229, 110}
\definecolor{r_pillow}{RGB}{255, 187, 120}
\definecolor{r_bench}{RGB}{141, 91, 229}
\definecolor{r_tv-screen}{RGB}{112, 128, 144}
\definecolor{r_cabinet}{RGB}{196, 156, 148}
\definecolor{r_pillar}{RGB}{197, 176, 213}
\definecolor{r_blanket}{RGB}{44, 160, 44}
\definecolor{r_tv-stand}{RGB}{148, 103, 189}
\definecolor{r_cushion}{RGB}{229, 91, 223}
\definecolor{r_bin}{RGB}{219, 219, 141}
\definecolor{r_vent}{RGB}{192, 229, 91}
\definecolor{r_bed}{RGB}{88, 218, 137}
\definecolor{r_stool}{RGB}{58, 98, 137}
\definecolor{r_picture}{RGB}{177, 82, 239}
\definecolor{r_indoor-plant}{RGB}{255, 127, 14}
\definecolor{r_desk}{RGB}{237, 204, 37}
\definecolor{r_comforter}{RGB}{41, 206, 32}
\definecolor{r_nightstand}{RGB}{62, 143, 148}
\definecolor{r_shelf}{RGB}{34, 14, 130}
\definecolor{r_vase}{RGB}{143, 45, 115}
\definecolor{r_plant-stand}{RGB}{137, 63, 14}
\definecolor{r_basket}{RGB}{23, 190, 207}
\definecolor{r_plate}{RGB}{16, 212, 139}
\definecolor{r_monitor}{RGB}{90, 119, 201}
\definecolor{r_pipe}{RGB}{125, 30, 141}
\definecolor{r_panel}{RGB}{150, 53, 56}
\definecolor{r_desk-organizer}{RGB}{186, 197, 62}
\definecolor{r_wall-plug}{RGB}{227, 119, 194}
\definecolor{r_book}{RGB}{38, 100, 128}
\definecolor{r_box}{RGB}{120, 31, 243}
\definecolor{r_clock}{RGB}{154, 59, 103}
\definecolor{r_sculpture}{RGB}{169, 137, 78}
\definecolor{r_tissue-paper}{RGB}{143, 245, 111}
\definecolor{r_camera}{RGB}{37, 230, 205}
\definecolor{r_tablet}{RGB}{14, 16, 155}
\definecolor{r_pot}{RGB}{208, 49, 84}
\definecolor{r_bottle}{RGB}{237, 80, 38}
\definecolor{r_candle}{RGB}{138, 175, 62}
\definecolor{r_bowl}{RGB}{158, 218, 229}
\definecolor{r_cloth}{RGB}{38, 96, 167}
\definecolor{r_switch}{RGB}{190, 77, 246}

\definecolor{s_wall}{RGB}{196, 51, 182}
\definecolor{s_floor}{RGB}{174, 199, 232}
\definecolor{s_cabinet}{RGB}{188, 189, 34}
\definecolor{s_bed}{RGB}{152, 223, 138}
\definecolor{s_chair}{RGB}{255, 152, 150}
\definecolor{s_sofa}{RGB}{214, 39, 40}
\definecolor{s_table}{RGB}{91, 135, 229}
\definecolor{s_door}{RGB}{31, 119, 180}
\definecolor{s_window}{RGB}{229, 91, 104}
\definecolor{s_bookshelf}{RGB}{247, 182, 210}
\definecolor{s_picture}{RGB}{91, 229, 110}
\definecolor{s_counter}{RGB}{255, 187, 120}
\definecolor{s_desk}{RGB}{141, 91, 229}
\definecolor{s_curtain}{RGB}{112, 128, 144}
\definecolor{s_refrigerator}{RGB}{196, 156, 148}
\definecolor{s_shower curtain}{RGB}{197, 176, 213}
\definecolor{s_toilet}{RGB}{44, 160, 44}
\definecolor{s_sink}{RGB}{148, 103, 189}
\definecolor{s_bathtub}{RGB}{229, 91, 223}
\definecolor{s_furniture}{RGB}{219, 219, 141}
\definecolor{s_background}{RGB}{192, 229, 91}

\definecolor{spp_wall}{RGB}{196, 51, 182}
\definecolor{spp_ceiling}{RGB}{174, 199, 232}
\definecolor{spp_floor}{RGB}{188, 189, 34}
\definecolor{spp_table}{RGB}{152, 223, 138}
\definecolor{spp_door}{RGB}{255, 152, 150}
\definecolor{spp_ceiling lamp}{RGB}{214, 39, 40}
\definecolor{spp_cabinet}{RGB}{91, 135, 229}
\definecolor{spp_blinds}{RGB}{31, 119, 180}
\definecolor{spp_curtain}{RGB}{229, 91, 104}
\definecolor{spp_chair}{RGB}{247, 182, 210}
\definecolor{spp_storage cabinet}{RGB}{91, 229, 110}
\definecolor{spp_office chair}{RGB}{255, 187, 120}
\definecolor{spp_bookshelf}{RGB}{141, 91, 229}
\definecolor{spp_whiteboard}{RGB}{112, 128, 144}
\definecolor{spp_window}{RGB}{196, 156, 148}
\definecolor{spp_box}{RGB}{197, 176, 213}
\definecolor{spp_window frame}{RGB}{44, 160, 44}
\definecolor{spp_monitor}{RGB}{148, 103, 189}
\definecolor{spp_shelf}{RGB}{229, 91, 223}
\definecolor{spp_doorframe}{RGB}{219, 219, 141}
\definecolor{spp_pipe}{RGB}{192, 229, 91}
\definecolor{spp_heater}{RGB}{88, 218, 137}
\definecolor{spp_kitchen cabinet}{RGB}{58, 98, 137}
\definecolor{spp_sofa}{RGB}{177, 82, 239}
\definecolor{spp_windowsill}{RGB}{255, 127, 14}
\definecolor{spp_bed}{RGB}{237, 204, 37}
\definecolor{spp_shower wall}{RGB}{41, 206, 32}
\definecolor{spp_trash can}{RGB}{62, 143, 148}
\definecolor{spp_book}{RGB}{34, 14, 130}
\definecolor{spp_plant}{RGB}{143, 45, 115}
\definecolor{spp_blanket}{RGB}{137, 63, 14}
\definecolor{spp_tv}{RGB}{23, 190, 207}
\definecolor{spp_computer tower}{RGB}{16, 212, 139}
\definecolor{spp_kitchen counter}{RGB}{90, 119, 201}
\definecolor{spp_refrigerator}{RGB}{125, 30, 141}
\definecolor{spp_jacket}{RGB}{150, 53, 56}
\definecolor{spp_electrical duct}{RGB}{186, 197, 62}
\definecolor{spp_sink}{RGB}{227, 119, 194}
\definecolor{spp_bag}{RGB}{38, 100, 128}
\definecolor{spp_picture}{RGB}{120, 31, 243}
\definecolor{spp_pillow}{RGB}{154, 59, 103}
\definecolor{spp_towel}{RGB}{169, 137, 78}
\definecolor{spp_suitcase}{RGB}{143, 245, 111}
\definecolor{spp_backpack}{RGB}{37, 230, 205}
\definecolor{spp_crate}{RGB}{14, 16, 155}
\definecolor{spp_keyboard}{RGB}{208, 49, 84}
\definecolor{spp_rack}{RGB}{237, 80, 38}
\definecolor{spp_toilet}{RGB}{138, 175, 62}
\definecolor{spp_paper}{RGB}{158, 218, 229}
\definecolor{spp_printer}{RGB}{38, 96, 167}
\definecolor{spp_poster}{RGB}{190, 77, 246}
\definecolor{spp_painting}{RGB}{0, 0, 0}
\definecolor{spp_microwave}{RGB}{208, 193, 72}
\definecolor{spp_board}{RGB}{55, 220, 57}
\definecolor{spp_shoes}{RGB}{10, 125, 140}
\definecolor{spp_socket}{RGB}{76, 38, 202}
\definecolor{spp_bottle}{RGB}{191, 28, 135}
\definecolor{spp_bucket}{RGB}{211, 120, 42}
\definecolor{spp_cushion}{RGB}{118, 174, 76}
\definecolor{spp_basket}{RGB}{17, 242, 171}
\definecolor{spp_shoe rack}{RGB}{20, 65, 247}
\definecolor{spp_telephone}{RGB}{208, 61, 222}
\definecolor{spp_file folder}{RGB}{162, 62, 60}
\definecolor{spp_cloth}{RGB}{210, 235, 62}
\definecolor{spp_blind rail}{RGB}{45, 152, 72}
\definecolor{spp_laptop}{RGB}{35, 107, 149}
\definecolor{spp_plant pot}{RGB}{160, 89, 237}
\definecolor{spp_exhaust fan}{RGB}{227, 56, 125}
\definecolor{spp_cup}{RGB}{169, 143, 81}
\definecolor{spp_coat hanger}{RGB}{42, 143, 20}
\definecolor{spp_light switch}{RGB}{25, 160, 151}
\definecolor{spp_speaker}{RGB}{82, 75, 227}
\definecolor{spp_table lamp}{RGB}{253, 59, 222}
\definecolor{spp_air vent}{RGB}{240, 130, 89}
\definecolor{spp_clothes hanger}{RGB}{123, 172, 47}
\definecolor{spp_kettle}{RGB}{71, 194, 133}
\definecolor{spp_smoke detector}{RGB}{24, 94, 205}
\definecolor{spp_container}{RGB}{134, 16, 179}
\definecolor{spp_power strip}{RGB}{159, 32, 52}
\definecolor{spp_slippers}{RGB}{213, 208, 88}
\definecolor{spp_paper bag}{RGB}{64, 158, 70}
\definecolor{spp_mouse}{RGB}{18, 163, 194}
\definecolor{spp_cutting board}{RGB}{65, 29, 153}
\definecolor{spp_toilet paper}{RGB}{177, 10, 109}
\definecolor{spp_paper towel}{RGB}{152, 83, 7}
\definecolor{spp_pot}{RGB}{83, 175, 30}
\definecolor{spp_clock}{RGB}{18, 199, 153}
\definecolor{spp_pan}{RGB}{61, 81, 208}
\definecolor{spp_tap}{RGB}{213, 85, 216}
\definecolor{spp_jar}{RGB}{170, 53, 42}
\definecolor{spp_soap dispenser}{RGB}{161, 192, 38}
\definecolor{spp_binder}{RGB}{23, 241, 91}
\definecolor{spp_bowl}{RGB}{12, 103, 170}
\definecolor{spp_tissue box}{RGB}{151, 41, 245}
\definecolor{spp_whiteboard eraser}{RGB}{133, 51, 80}
\definecolor{spp_toilet brush}{RGB}{184, 162, 91}
\definecolor{spp_spray bottle}{RGB}{50, 138, 38}
\definecolor{spp_headphones}{RGB}{31, 237, 236}
\definecolor{spp_stapler}{RGB}{39, 19, 208}
\definecolor{spp_marker}{RGB}{223, 27, 180}
\usepackage{balance}
\usepackage{soul}

\def\ours{OVO\xspace}
\def\clipours{CLIP merging\xspace}
\begin{document}
\author{Tomas Berriel Martins\\
University of Zaragoza
\and
Martin R. Oswald\\
University of Amsterdam\\
\and
Javier Civera\\
University of Zaragoza}
\title{Open-Vocabulary Online Semantic Mapping for SLAM}

\maketitle
\begin{abstract}
This paper presents an \underline{O}pen-\underline{V}ocabulary \underline{O}nline 3D semantic mapping pipeline, that we denote by its acronym \ours. 
Given a sequence of posed RGB-D frames, we detect and track 3D segments, which we describe using CLIP vectors. These are computed from the viewpoints where they are observed by a novel CLIP merging method. Notably, our \ours has a significantly lower computational and memory footprint than offline baselines, while also showing better segmentation metrics than offline and online ones. Along with superior segmentation performance, we also show experimental results of our mapping contributions integrated with two different full SLAM backbones (Gaussian-SLAM and ORB-SLAM2), being the first ones using a neural network to merge CLIP descriptors and demonstrating end-to-end open-vocabulary online 3D mapping with loop closure.
\end{abstract}
    
\section{Introduction}
\label{sec:intro}
Semantic mapping targets the estimation of  the category to which each element in a scene belongs, along with a consistent geometric representation. Rich semantic representations in 3D are essential for advanced robotic applications.
Traditionally, semantic 3D reconstruction has relied on a closed-set approach in both offline~\cite{bao2011semantic,kundu2022panoptic} and online~\cite{mccormac2017semanticfusion,rosinol2020kimera} settings, including integrations into semantic Simultaneous Localization and Mapping (SLAM) systems~\cite{civera2011towards,zhu2024sni,li2024sgs}.
However, these methods are constrained by a predefined set of categories, which limits their flexibility and applicability in open-ended, real-world environments.

\begin{figure}[ht!]
    \centering
    \includegraphics[width=0.9\linewidth]{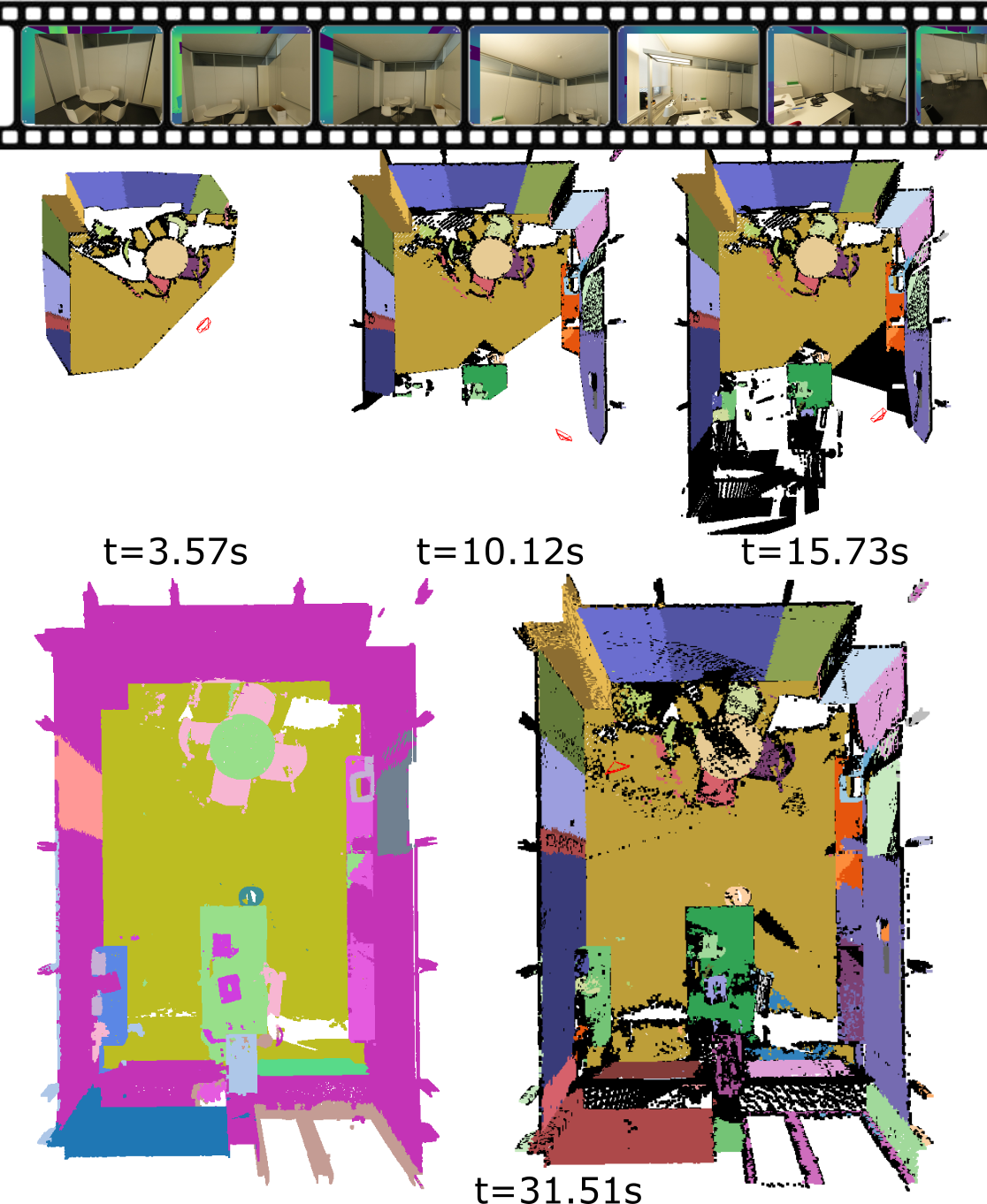}
    \resizebox{\linewidth}{!}
{
\begin{tabular}{c}
\ColorMapCircle{spp_ceiling lamp}\,ceiling lamp
\ColorMapCircle{spp_bottle}\,bottle
\ColorMapCircle{spp_telephone}\,telephone 
\ColorMapCircle{spp_shelf}\,shelf
\small\ColorMapCircle{spp_wall}\,wall
\ColorMapCircle{spp_chair}\,chair
\ColorMapCircle{spp_door}\,door
\ColorMapCircle{spp_box}\,box\\
\small\ColorMapCircle{spp_whiteboard}\,whiteboard
\ColorMapCircle{spp_ceiling}\,ceiling
\ColorMapCircle{spp_cabinet}\,cabinet
\ColorMapCircle{spp_blinds}\,blinds 
\ColorMapCircle{spp_socket}\,socket 
\ColorMapCircle{spp_heater}\,heater
\ColorMapCircle{spp_table}\,table
\ColorMapCircle{spp_floor}\,floor
\ColorMapCircle{spp_window}\,window
\end{tabular}
}
    \caption{\textbf{\ours mapping.} Given a RGB-D set of keyframes (\textbf{top}), our method successively reconstructs a 3D open-vocabulary representation of a scene over time (\textbf{middle}). 
    At any moment, both semantic labels (\textbf{bottom left}) as well as instance labels (\textbf{bottom right}) can be effectively recovered.
    }
    \label{fig:teaser}
\end{figure}

Following the emergence of Contrastive Language-Image Pre-training (CLIP)~\cite{radford2021clip}, there has been a surge of interest in open-vocabulary 3D representations \cite{Peng2023OpenScene,takmaz2023openmask3d,nguyen2024open3dis}, including efforts in online mapping \cite{conceptfusion, gu2024conceptgraphs, yamazaki2024open}—though not yet in full SLAM systems.
While these recent approaches have shown strong performance, their dependence on offline processing or ground-truth camera poses for mapping significantly limits their applicability in robotics, augmented reality, and virtual reality scenarios.

In this paper, we present \ours, an \underline{O}pen-\underline{V}ocabulary \underline{O}nline mapping algorithm, which we integrate into two distinct visual SLAM pipelines. An example of our online reconstruction results is shown in \cref{fig:teaser}.
Our method processes RGB-D keyframes to generate 3D segments, each associated with a CLIP embedding.
These segments are initialized by back-projecting masks predicted by Segment Anything Model (SAM) 2.1~\cite{ravi2024sam2}, and are tracked over time by projecting them into 2D and matching against new masks.
Each 3D segment's CLIP descriptor is selected from the keyframe views with the best visibility.
Additionally, we introduce a novel model to extract per-instance CLIP descriptors directly from images, which are then assigned to the corresponding 3D masks.
{Our \clipours employs a neural network that learns per-dimension weighting to fuse CLIP descriptors of the same instance, while effectively generalizing to unseen classes and environments.}
Our pipeline not only operates online and supports loop-closure optimization, but also outperforms existing baselines in segmentation accuracy.
\section{Related Work}
\label{sec:related}
Our \ours estimates consistent 3D open-vocabulary semantics and seamlessly integrates with SLAM pipelines. Unlike previous methods that either use a closed set of categories, offline processing, 2D semantic representations or odometry. \cref{tab:sota} provides a comparative summary of recent related works based on these aspects, with further details discussed in the remainder of this section.

\boldparagraph{Open-Vocabulary Image Semantics.} 
The introduction of Contrastive Language-Image Pretraining (CLIP)~\cite{radford2021clip}, which encodes image and text tokens into a shared latent space, revolutionized semantic segmentation.
By computing similarity to text inputs, CLIP enables classification into any category expressible in language.
Several variations of CLIP have enhanced its performance~\cite{zhai2023siglip,  cherti2023openclip} and improved feature granularity, aiming to generate dense feature vectors~\cite{zhou2022extract, sun2024alpha} rather than per-image representations. 
While closed-vocabulary methods outperform on predefined sets, open-vocabulary offers optimization-free generalization, highly relevant for diverse applications.
%

\boldparagraph{Offline 3D Open-Vocabulary from 3D point clouds.} Most open-vocabulary 3D semantic approaches assume a known 3D point cloud.
OpenScene~\cite{Peng2023OpenScene} leverages OpenSeg~\cite{ghiasi2022scaling} to compute CLIP features from images and trains a network to associate 2D pixels with 3D points.
For each 3D point it performs average pooling on CLIP vectors from multiple views and supervises an encoder to directly assign CLIP features to 3D point clouds. 
OpenMask3D~\cite{takmaz2023openmask3d} selects $k$ views per object, crops its 2D SAM mask to compute a CLIP features, and then features are average-pooled across crops and views. 
Open3DIS~\cite{nguyen2024open3dis} integrates SuperPoint~\cite{detone2018superpoint} with 2D instance segmentations and a 3D instance segmentator to generate multiple 3D instance proposals, describing each with CLIP features following OpenMask3D~\cite{takmaz2023openmask3d}.
In contrast, OpenYolo-3D~\cite{boudjoghra2024openyolo} uses a 2D open-vocabulary object detector instead of relying on 2D instance masks and CLIP features.
It classifies each object based on the most common class across all views.
While this approach eliminates the need for CLIP feature extraction, it limits each scene to a predefined set of classes.

\boldparagraph{Offline 3D Open-Vocabulary from RGB and RGB-D.} OpenNeRF~\cite{engelmann2024opennerf} optimizes a NeRF to encode the scene representation along with per-pixel CLIP features from OpenSeg. The OpenSeg features are projected into 3D to compute the mean and covariance of 3D points. The NeRF then renders novel views, prioritizing areas with high covariance to compute additional OpenSeg features and refine the model.
Hierarchical Open-Vocabulary 3D Scene Graphs (HOV-SG) \cite{hovsg} relies on an offline hierarchical global fusion approach that requires precomputing 3D segments and features for all frames. 
These 3D segments and features are incrementally fused by merging observations across consecutive frames.
The authors argue that relying solely on masked segments, as in Concept-Graphs~\cite{gu2024conceptgraphs}, discards crucial contextual information. To address this, they propose a descriptor that merges in a handcrafted manner three CLIP embeddings per mask: (1) the full image, (2) the masked segment without background, and (3) the masked segment with background. We adopt this strategy, and contribute by proposing a novel approach to learn the CLIP merging operation.

\boldparagraph{Online Semantics.} To date, online semantic methods have focused mostly  on closed vocabularies. SemanticFusion~\cite{mccormac2017semanticfusion} was one of the first semantic SLAM pipelines, predicting per-pixel closed-set categories and fusing predictions from different views in 3D space. Fusion++~\cite{mccormac2018fusion++} uses Mask-RCNN~\cite{he2017mask} to initialize per-object Truncated Signed Distance Functions (TSDFs), building a persistent object-graph representation.
In contrast, PanopticFusion~\cite{narita2019panopticfusion} combines predicted instances and class labels (including background) to generate pixel-wise panoptic predictions, which are then integrated into a 3D mesh. More recent works, such as those by Menini et al.~\cite{menini2021real} and ALSTER~\cite{weder2023alster}, jointly reconstruct geometry and semantics in a SLAM framework.
Additionally, NIS-SLAM~\cite{zhai2024nis} trains a multi-resolution tetrahedron NeRF to encode color, depth and semantics. 
NEDS-SLAM~\cite{ji2024neds} is a 3DGS-based SLAM system with embedded semantic features to learn an additional semantic representation of a closed set of classes.
Similarly, Hi-SLAM~\cite{li2024hi} and SGS-SLAM~\cite{li2024sgs} augment a 3DGS SLAM with semantic ids of predefined set of classes. 
These approaches either assume known 2D ground-truth closed set of semantic classes (and therefore only tackle a multi-view fusion problem), or only represent 2D semantics, with limited capabilities for 3D segmentation or precise 3D object localization.
More recently, OpenFusion~\cite{yamazaki2024open} and Concept-Graphs~\cite{gu2024conceptgraphs} integrated open-vocabulary semantic descriptors into online 3D mapping pipelines. Concept-Graphs relies on the naive mask-cropping to compute CLIP descriptors, while OpenFusion uses SEEM~\cite{zou2023segment} and creates a TSDF with 3D segments. 
None of them, however, addresses the integration into a full SLAM pipeline with loop closure optimization as we do. 

\begin{table}[t]
  \scriptsize
  \caption{Overview of 3D semantic reconstruction baselines.}
  \label{tab:sota}
  \centering
  \setlength{\tabcolsep}{8pt} 
  \begin{tabular}{l c c c c }
    \toprule
     \multirow{2}{*}{Method}      & Open     & 3D   &  \multirow{2}{*}{Online}         & Loop \\
     & Vocabulary & semantics  & & Closure \\
    \midrule
    OpenScene~\cite{Peng2023OpenScene}     & \cmark & \cmark & \xmark  & -  \\
    OpenMask3D~\cite{takmaz2023openmask3d} & \cmark & \cmark & \xmark  & - \\
    Open3DIS~\cite{nguyen2024open3dis}     & \cmark & \cmark & \xmark  & -  \\
    HOV-SG~\cite{hovsg}                    & \cmark & \cmark & \xmark & -  \\
    OpenNeRF~\cite{engelmann2024opennerf}  & \cmark & \cmark & \xmark  & -  \\
    \hdashline \noalign{\vskip 1pt}
    NEDS-SLAM~\cite{ji2024neds}            & \xmark & \xmark & \cmark  & \xmark \\
    NIS-SLAM~\cite{zhai2024nis}            & \xmark & \xmark & \cmark  & \xmark  \\
    SGS-SLAM~\cite{li2024sgs}            & \xmark & \cmark & \cmark  & \xmark  \\
    Kimera-VIO~\cite{rosinol2020kimera}    & \xmark & \cmark & \cmark & \xmark  \\ \hdashline
    Concept-Fusion~\cite{conceptfusion}        & \cmark & \cmark & \cmark  & \xmark  \\
    Concept-Graphs~\cite{gu2024conceptgraphs}        & \cmark & \cmark & \cmark  & \xmark  \\
    Open-Fusion~\cite{yamazaki2024open}        & \cmark & \cmark & \cmark  & \xmark  \\
    \hdashline \noalign{\vskip 1pt}
    \textbf{\ours (ours)}             & \cmark & \cmark & \cmark & \cmark \\
    \bottomrule
  \end{tabular}
\end{table}
\begin{figure*}[ht!]
    \centering
    \includegraphics[width=\linewidth]{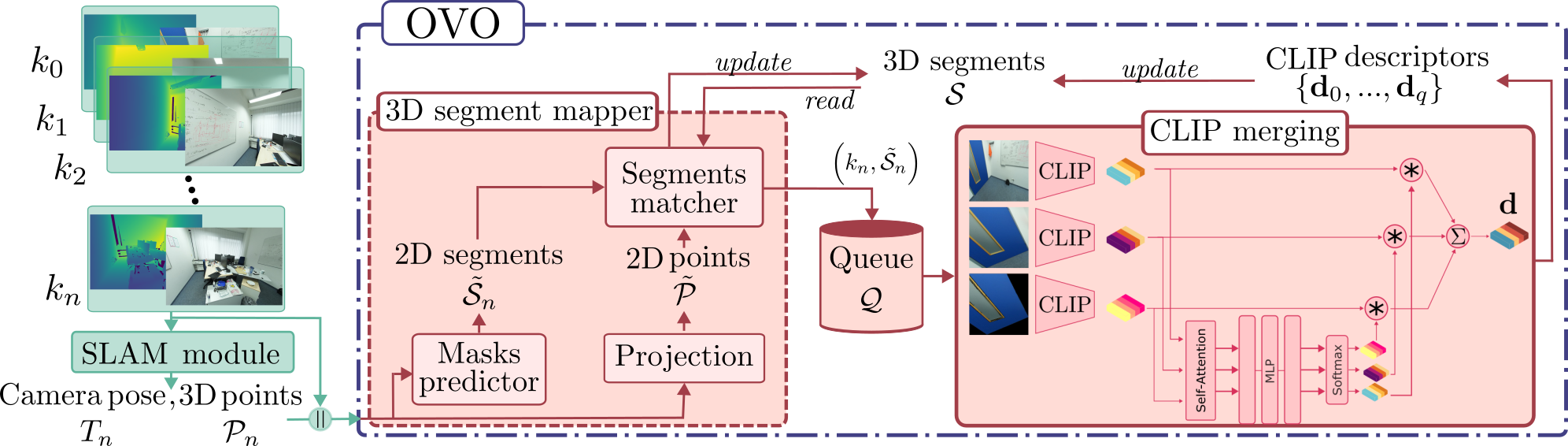}
    \caption{\textbf{Overview.} From a stream of RGB-D keyframes, \ours builds, online, a 3D semantic representation of the scene. It relies on a 3D segment mapper to cluster 3D points into 3D segments; a queue to distribute the CLIP extraction computation, and a novel CLIP merging method to aggregate CLIP descriptors from multiple keyframes into one for each 3D segment.}
    \label{fig:ovo}
\end{figure*}
\section{\ours Methodology}
~\ours relies on a parallel-tracking-and-mapping architecture, as first defined by Klein and Murray~\cite{klein2007parallel} and adopted by most visual SLAM implementations~\cite{campos2021orb}. \cref{fig:ovo} shows an overview of \ours.
It takes as input a stream of RGB-D keyframes ($\{ k_0, \hdots, k_n\}$ in the figure) and their respective poses and local point clouds. 
From this 3D representation, \cref{sec:map}, \ours extracts and tracks a set of 3D segments covering the whole representation (\emph{3D segment mapper} in the figure, detailed in~\cref{sec:mapper}). 
We compute a CLIP descriptor per each segment's viewpoint merging 3 different CLIPs (\emph{CLIP merging} in the figure, detailed in~\cref{sec:merging}). Then assign to the 3D segment the most representative descriptor, \cref{sec:sem_computation}. 
When the SLAM module performs a loop closure or bundle-adjustment optimization, a routine searches for repeated 3D segments, and fuses those that were not correctly tracked,~\cref{sec:loop_closure}.
\subsection{Map Definition}
\label{sec:map}
%
Its input is an RGB-D video \(\mathcal{V} = \{f_0, \hdots, f_\tau\}\), \(f_\tau \in \mathbb{N}_{\leq 255}^{w \times h \times 3} \times \mathbb{R}_{>0}\) representing the RGB-D frame of size $w \times h$ captured at time step \(\tau\). A SLAM front-end estimates in real-time the pose \(T_n\) of every frame \(f_\tau\) in the world reference frame. The SLAM back-end selects a set of keyframes \(\mathcal{K} = \{k_0, \hdots\, k_n\} \subset \mathcal{V}\) from which it iteratively refines their poses $\mathcal{T} = \{T_{0}, \hdots, T_{n}\}$, \(T_{n} \in SE(3)\) asynchronously, at a rate lower than the video rate of the tracking thread. 

Our scene representation or `map' $\mathcal{M} = \{ \mathcal{T}, \mathcal{P}, \mathcal{S} \}$, consists on these keyframe poses $\mathcal{T}$, a point cloud $\mathcal{P} = \{P_0, \hdots, P_m\}$ and a set of 3D segments $\mathcal{S} = \{S_0, \hdots, S_q\}$, being $q$ the identifier of the last added segment. 
Every map point $P = \big( \begin{bmatrix}
x & y & z
\end{bmatrix}^\top, \ l_p \big)$ is defined by its 3D coordinates $\begin{bmatrix}
x & y & z 
\end{bmatrix}^\top \in \mathbb{R}^3$ and a discrete label $l_p \in \{-1, 0, 1, \hdots, q\}$, $l_p>-1$ indicating the 3D segment the point belongs to, and $l_p=-1$ indicating that it is unassigned. 
The dense point cloud \(\mathcal{P}\) is built concatenating at each keyframe \(k_n\) the estimated 3D points \(\mathcal{P}_n\) provided by the SLAM front-end. 
If the SLAM front-end does not estimate a dense point cloud, \(\mathcal{P}_n\) is computed as the unprojection of the input depth map to 3D using the estimated camera pose \(T_{n} \in SE(3)\). To avoid \(\mathcal{P}\) growing unconstrained, a pixel is not projected to 3D if a previously unoccluded 3D point falls inside its neighborhood when projected back to 2D. 
For every 3D point, occlusion is assessed by comparing its projected depth to its measured depth in the 2D pixel it is projected.
Every 3D segment $S = \left( \mathbf{d}, \kappa \right)$ has a unique identifier, its semantics are described by a CLIP feature $\mathbf{d} \in \mathbb{R}^d$, and stores in a heap $\kappa$ the indices of the best keyframes in which $S$ was seen, ordered by visibility scores.
\subsection{3D Segment Mapper}
\label{sec:mapper}
\begin{algorithm}[t]
\normalsize 
\caption{3D Segment Mapper} 
\begin{algorithmic}[1] 
\footnotesize
\Function{3D\_segment\_mapper}{$\mathcal{P}$, $\mathcal{S}$, $k_n, T_n$}

    \State $\tilde{\mathcal{S}}_n\leftarrow$ segment\_keyframe$(k_n)$
    \State $\tilde{\mathcal{P}}_n \leftarrow$ project\_point\_cloud$(\mathcal{P},T_n)$
    \For{$\left(s, l_s\right) $ in $\tilde{\mathcal{S}}_n$} \Comment{For every 2D segment in $k_n$}
    \State $mode, v \leftarrow$ get\_label\_mode\_and\_votes$(\tilde{\mathcal{P}}_n,s,\epsilon)$
    \If{$v > \epsilon$} \Comment{\#votes greater than threshold}
    \If{$mode = -1$}
    \State $S_{q+1} \leftarrow$ new\_3D\_segment$(q+1, n, s)$
    \State $\mathcal{S} \leftarrow \mathcal{S} \cup\{S_{q+1}\}$
    \State $l_s \leftarrow q+1$
    \Else
    \State $\mathcal{S} \leftarrow$ update\_3D\_segment$(S_{mode}, n, s)$
    \State $l_s \leftarrow z_l$
    \EndIf
    \EndIf
    \EndFor
    \State $\tilde{\mathcal{S}}_n \leftarrow$ merge\_and\_prune\_2D\_segments$(\tilde{\mathcal{S}}_n)$
    \State $\mathcal{P} \leftarrow$ update\_pcd\_labels$(\mathcal{P},\tilde{\mathcal{P}}_n, \tilde{\mathcal{S}}_n)$
    \State \Return $\mathcal{P}$, $\mathcal{S}, \tilde{\mathcal{S}}_n$
\EndFunction
\end{algorithmic}
\label{alg:3dsegmentor}
\end{algorithm}
\begin{figure*}[tb]
    \centering
\includegraphics[width=\linewidth]{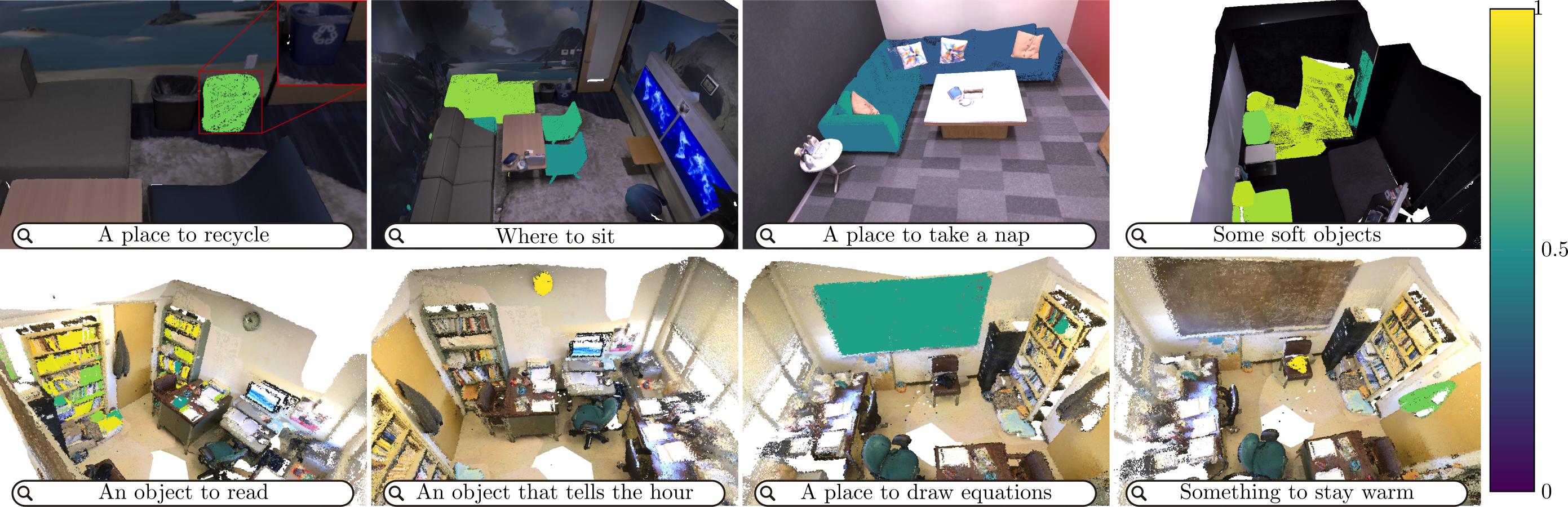}
    \caption{\textbf{Out-of-distribution queries.} From left to right, top to bottom, observe how common-language queries allow to differentiate bins based on a recycling symbol; recongize sofas and chairs as places to sit; that you can take a nap in a sofa, pillows and couches are soft objects, and books are readable, that the clock tells the hour, the blackboard is to draw equations, and the jacket is something to stay warm. Colorbar shows similarity strength. 
    }
    \label{fig:gen_queries}    
\end{figure*}
For every new keyframe $k_n$, we run an image segmentation model that returns a set of 2D segments $\tilde{\mathcal{S}_n}=\{ \left(s_0,l_{s0}\right), \left(s_1,l_{s1}\right), \hdots\}$, each segment being composed of a mask $s$ and a label $l_s$, which is initialized as $l_s:=-1$. 
We then select the 3D map points in $k_n$'s frustum, project them to $k_n$, and remove occluded points by comparing their projected depth to the input depth. 
In this manner, we obtain the 2D point set $\tilde{\mathcal{P}}_n=\{p_0, p_1, \hdots\}$, for which $p = \big( \begin{bmatrix}
u & v
\end{bmatrix}^\top, \ l_p \big)$.
We compute the label mode of all points $p$ within a segment $s$, that we will represent slightly abusing notation as $z_l := \arg\max_{l_p}\left(\tilde{\mathcal{P}}\cap s\right)$. 
If the mode receives less votes \(v\) than a predefined threshold $\epsilon$, we discard $s$.
If not, two possibilities can occur:
\begin{enumerate}
    \item If $z_l\!=\!-1$, we set $z_l\!:=\!q\!+\!1$ and initialize a new 3D segment $S_{q+1}$ with an empty CLIP feature $\mathbf{d}$ (filled later as described in Sec.~\ref{sec:sem_computation}), and a keyframe heap $\kappa\!:=\!\{\left(n, r\right)\}$, initialized with $k_n$'s index and $s$' visibility score $r$.
    \item Otherwise, 2D segment $s$ is a match for 3D segment $S_{z_l}$ and the keyframe will be inserted into \(\kappa\), and stored if it is one of the best views or if \(\kappa\) is not full.
\end{enumerate}
For both, the unassigned 3D points and 2D segment's labels, $l_p$ and  $l_s$ are updated to the identifier of the matched ${S}_{z_l}$.

After matching all 2D masks, those that share the same $l_s$ are merged.
Finally, once all masks are gathered in $\tilde{\mathcal{S}}_n$, the tuple $\big(k_n,\tilde{\mathcal{S}}_n\big)$ is pushed to the queue \(\mathcal{Q}\).
Keyframes and masks remain in $\mathcal{Q}$ until processing resources become available to compute the CLIP descriptors for the highest-scoring 2D segments. 
%
%
\subsection{Loop Closure}
\label{sec:loop_closure}
%
When the SLAM module closes a loop or completes a Global Bundle Adjustment, \ours updates both its map and the set of 3D instances. We denote both after the update as \(\mathcal{M}^\prime\) and \(\mathcal{S}^\prime\).
For each updated keyframe \(T_n^\prime\in\mathcal{T}^\prime\), its associated local point cloud is also updated by propagating the pose correction as \(
    \mathcal{P}_{n} := T_{n}^\prime \ T_{n}^{-1} \ \mathcal{P}_n
\). 
This transforms the points from the world frame to the original keyframe's and back using the updated pose \(T_{n}^\prime\).
Keyframes that are removed during SLAM optimization are discarded along with their associated 3D points.
After updating the 3D points, the temporary queue \(\mathcal{Q}\) is cleared. Next, the set of 3D instances \(\mathcal{S}^\prime\) is pruned by removing instances whose associated points were entirely deleted during optimization.
Following, instance fusion is performed by comparing remaining pairs of 3D instances. Two instances are merged if they satisfy the following criteria: \textbf{(1)} The distance between their point cloud centroids is \(<150\)cm, \textbf{(2)} the cosine similarity between their CLIPs is \(>0.8\), and \textbf{(3)} more than \(50\%\) of their points lie within $10$cm of a point in the other instance.
For a pair of segments \(S_i\) and \(S_j\) to be merged, their point indices are unified as \( \kappa_i := \kappa_i \cup \kappa_j \), and all map points previously labeled as \(j\) are reassigned to \(i\), i.e., 
\(\forall P_k \in \mathcal{P} | l_k = j ,\implies l_k := i\).
%
\subsection{CLIP Descriptors
} \label{sec:sem_computation}
%
When a tuple $\big(k_q,\tilde{\mathcal{S}}_q\big)$ is popped from $\mathcal{Q}$, only the matched 2D segments for which $k_q$ is still in the $\kappa$ of their 3D instance $S$ are selected. 
A CLIP descriptor $\mathbf{d}$ is computed for each of them as explained in \cref{sec:merging}. 
Then, the final descriptor for a 3D segment $S$ is selected between the 2D segments in its keyframes' heap $\kappa$, as the CLIP descriptor with the smallest aggregated distance to the rest.
To query the 3D semantic representation, text queries are encoded to CLIP space.
 Then, we compute the cosine similarity between the CLIP descriptor of the query and the descriptor $\mathbf{d}$ of each 3D segment in $\mathcal{S}$. 
\begin{figure*}[tb]
\centering
\includegraphics[width=\textwidth]{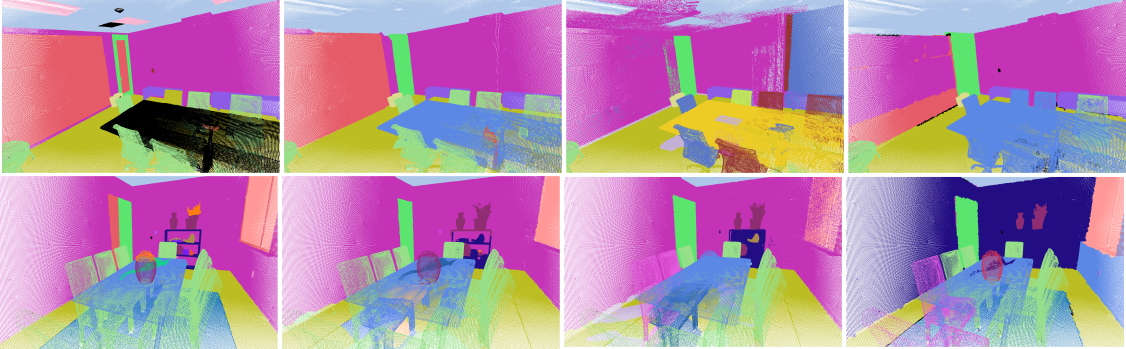}
\resizebox{\textwidth}{!}
{
\begin{tabular}{cccc}
        \vspace{-10px}\\
        \textit{\small \color{black} Ground truth} &
        \textit{\small \color{black} \ours -Gaussian-SLAM (ours)} &
        \textit{\small \color{black} HOV-SG~\cite{hovsg}} &
        \textit{\small \color{black} Open3DIS~\cite{nguyen2024open3dis}}\\
        \hspace{0.22\textwidth} &
        \hspace{0.22\textwidth} &
        \hspace{0.22\textwidth} &
        \hspace{0.22\textwidth}
    \end{tabular}
}
\resizebox{\textwidth}{!}{
\begin{tabular}{c}
\vspace{-22px}
\ColorMapCircle{r_panel}\,panel
\ColorMapCircle{r_vase}\,vase
\ColorMapCircle{r_clock}\,clock
\ColorMapCircle{r_pot}\,pot 
\ColorMapCircle{r_window}\,window
\ColorMapCircle{r_bottle}\,bottle
\ColorMapCircle{r_indoor-plant}\,indoor-plant
\ColorMapCircle{r_blinds}\,blinds
\ColorMapCircle{r_lamp}\,lamp
\ColorMapCircle{r_wall-plug}\,wall-plug
\small\ColorMapCircle{r_wall}\,wall
\ColorMapCircle{r_switch}\,switch
\ColorMapCircle{r_bench}\,bench\\ \vspace{14px}
\small\ColorMapCircle{r_box}\,box 
\ColorMapCircle{r_shelf}\,shelf
\ColorMapCircle{r_stool}\,stool
\ColorMapCircle{r_rug}\,rug
\ColorMapCircle{r_table}\,table
\ColorMapCircle{r_ceiling}\,ceiling
\ColorMapCircle{r_bowl}\,bowl
\ColorMapCircle{r_camera}\,camera
\ColorMapCircle{r_door}\,door
\ColorMapCircle{r_plate}\,plate
\ColorMapCircle{r_chair}\,chair
\ColorMapCircle{r_vent}\,vent
\ColorMapCircle{r_bin}\,bin
\ColorMapCircle{r_desk}\,desk
\ColorMapCircle{r_floor}\,floor
\ColorMapCircle{r_sculpture}\,sculpture
\end{tabular}
}
\vspace{-14px}
\caption{\textbf{3D semantic segmentation on Replica.} \ours yields more accurate results in comparison to the two best offline baselines.}

\label{fig:replica}
\end{figure*}

\subsection{CLIP Merging}
\label{sec:merging}
Similarly to HOV-SG~\cite{hovsg}, for each 2D segment we compute three CLIP descriptors: 1) $\mathbf{d}_0$ for the full keyframe, 2) $\mathbf{d}_1$ for the segment masking the rest of the image out, and 3) $\mathbf{d}_2$ for the minimum bounding box that contains the segment.
In contrast, in our case, the CLIP descriptor $\mathbf{d}= \sum_{i=0}^2 \mathbf{w}_i \odot \mathbf{d}_i$ of a 2D segment is the result of merging the three descriptors $\mathbf{d}_{i=\{0, 1, 2\}}$ using a per-dimension weighted average with weights $\mathbf{w}_i \in \mathbb{R}^d$ ($\odot$ is the Hadamard product). Our weights $\mathbf{w}_{i=\{0, 1, 2\}}$ are predicted by a neural model, as shown in \cref{fig:ovo}. Note that HOV-SG's merging is done with hand-crafted scalar weights (\emph{i.e.}, $\mathbf{d}= \sum_{i=0}^2 {w}_i \mathbf{d}_i$, ${w}_i \in \mathbb{R}$).

As seen in 
\cref{fig:ovo}, the input to our CLIP merging  is three CLIPs $\mathbf{d}_{i=\{0, 1, 2\}}$. 
These are first passed by a transformer encoder, and the output is flattened and fed to a MLP, predicting the weights, and a softmax, forcing $\sum_{i=0}^2 \mathbf{w}_{i}=\mathbf{1}^d$.
Our \clipours is pre-trained following  SigLIP~\cite{zhai2023siglip}.
For a mini-batch \(\mathcal{B} = \left\{ (s_0, c_0), (s_1, c_1), \dots \right\}\) composed by pairs of 2D segments $s_j$ and semantic classes $c_j$, we minimize the sigmoid cosine similarity loss
\begin{equation}
   L = -\frac{1}{|\mathcal{B}|} \sum_{i=1}^{|\mathcal{B}|} \sum_{j=1}^{|\mathcal{B}|} \log\! \left( \frac{1}{1 \! +\! \exp(z_{ij}(-t \mathbf{d}_i \cdot \mathbf{y}_j + b))} \right)
\end{equation} 
between the merged CLIP descriptor \(\mathbf{d}_i\), and the CLIP embedding \(\mathbf{y}_j\) of the semantic class $c_j$ associated to the 2D segment $s_j$ in the same batch $\mathcal{B}$. \(z_{ij}\) is the label for a given image and class input, which equals \(1\) if they are paired and \(-1\) otherwise. \(b\) and \(t\) are learnable bias and temperature parameters, used to compensate the imbalance coming from negative pairs dominating the loss.
\section{Experiments}
\label{sec:experiments}
{First, we report \ours evaluation} on 3D online semantic mapping on two established datasets, one synthetic (Replica), and one real (ScanNetv2).
{Then, we present our \clipours evaluation} on semantic classification of images with ground-truth segmentation masks both on a dataset with multiple masks per image (ScanNet++) and with a single mask per image (ImageNet-S){, and against alternative methods integrated into \ours for 3D semantic mapping (Replica).}

{\boldparagraph{Implementation.} For \ours,} we implemented three different configurations {to show its flexibility}: (1)~\textbf{\ours-mapping}, that uses ground-truth camera poses, (2)~\textbf{\ours-Gaussian-SLAM}, {where we} integrate our contributions within Gaussian-SLAM~\cite{yugay2023gaussian}{, a SLAM method targeting novel-view synthesis and dense point cloud reconstruction, although not real-time}, and (3)~\textbf{\ours-ORB-SLAM2} for which we {integrate} {with ORB-SLAM2~\cite{mur2017orb}, a real-time feature-based SLAM system with loop-closure. While \ours-Gaussian-SLAM uses the center of 3D Gaussians as the dense point cloud \(\mathcal{P}\), for \textbf{\ours-ORB-SLAM2} we build a dense point cloud by registering the local point clouds from the RGB-D images}.
All three configurations use SAM2.1-l for 2D segmentation and SigLip ViT-SO400 for CLIP descriptors. 
%
{\textbf{Our \clipours}} has 5 self-attention layers with 8 heads, {a 1152 latent dimension, with drop-out of 0.1, }and 4 layers MLP {with \(3\times 1152\) input/output neurons and \(\times4\) inverse bottleneck with Leaky ReLU activations.} {It was trained with 4 Nvidia V100 GPUs, using Pytorch, with AdamW optimizer, learning rate \(1\times10^{-6}\), gradient clipping at 1, and batch size of 512 per GPU,} for 15 epochs using the top 100 semantic labels from ScanNet++ 250 training set. {To compensate for class imbalance, in the loss we weight each element of the batch by the inverse of their class frequency in the training set.}

\boldparagraph{Baselines.}
We evaluate~\clipours against baselines that compute local CLIP descriptors~\cite{conceptfusion, gu2024conceptgraphs,hovsg} (using all of them SigLIP-SO400M) and Alpha-CLIP~\cite{sun2024alpha}, a state-of-the-art model developed to condition CLIP using masks.
{Additionally, we include two variations of our \clipours trained in the same setup, in order to validate its design: directly predicting the fused descriptor, and predicting only one weight per descriptor.} 
As detailed in \cref{sec:related}, existing semantic SLAM pipelines do not construct a 3D representation that can be evaluated using 3D metrics for open-set classes. 
Thus, we compare \ours against similar 3D open-vocabulary online mapping systems, Concept-Graphs~\cite{gu2024conceptgraphs}, and OpenFusion~\cite{yamazaki2024open}; and the state-of-the-art 3D open-vocabulary offline baselines OpenScene~\cite{Peng2023OpenScene}, OpenNeRF~\cite{engelmann2024opennerf}, Open3DIS~\cite{nguyen2024open3dis} and HOV-SG~\cite{hovsg}.
Finally, we evaluate computational cost against Concept-Graphs, OpenFusion, HOV-SG and OpenNeRF, but exclude Open3DIS and OpenScene, as they rely on pre-processed 3D geometry and features.

\boldparagraph{Datasets.} 
ScanNet++~\cite{yeshwanthliu2023scannetpp} has 250 training and 50 validation indoor RGB+D scenes sequences. We use 2D rasterized masks for a total of 1.6M and 400k 2D instance samples respectively. Semantic classes are mapped into either the set of 100 most commons (used for training) or the full set of over 1.6k classes (used for evaluation).
ImageNet-S~\cite{gao2022luss} has a validation set of $\sim\!12$k images with 919 semantic labels. 
ScanNetv2~\cite{dai2017scannet} has a full validation set of 312 RGB+D sequences of real scenes (FVS). We also evaluate on the 5-scene subset used by HOV-SG (HVS). We use the original annotation set with 20 classes (ScanNet20) and the expanded set with 200 classes (ScanNet200~\cite{rozenberszki2022language}).
On Replica~\cite{replica19arxiv}, we use the standard 8-scene subset (\textit{office-0...4}, \textit{room-0...2}) and its 51 annotated classes.
\begin{table*}[ht]
\caption{3D semantic segmentation evaluation on Replica 51 classes, splitting by frequency tertiles: Head, Common and Tail.}
\label{tab:replica}
\centering
\setlength{\tabcolsep}{6.5pt}
\footnotesize
\begin{tabular}{l|ccc|cccccccc}
   \toprule
   & & \multirow[b]{2}{*}{\makecell{Geo-\\metry}}& \multirow[b]{2}{*}{\makecell{Camera pose / \\{ATE RMSE} {[cm]}}} & \multicolumn{2}{c}{All} & \multicolumn{2}{c}{Head} & \multicolumn{2}{c}{Common} & \multicolumn{2}{c}{Tail} \\
  \cmidrule(lr){5-6} \cmidrule(lr){7-8} \cmidrule(lr){9-10} \cmidrule(lr){11-12}
  Method & Online & &  & mIoU & mAcc & mIoU & mAcc & mIoU & mAcc & mIoU & mAcc \\
  \midrule
  OpenScene~\cite{Peng2023OpenScene} (Distilled)&\xmark& GT & GT & 14.8 & 23.0 & 30.2 & 41.1 & 12.8 & 21.3 & 1.4 & 6.7 \\
  OpenScene~\cite{Peng2023OpenScene} (Ensemble) &\xmark& GT & GT & 15.9 & 24.6 & 31.7 & 44.8 & 14.5 & 22.6 & 1.5 & 6.3 \\
  OpenNeRF~\cite{engelmann2024opennerf}         &\xmark  & Est.& GT & 20.4 & 31.7 & 35.4 & 46.2 & 20.1 & 31.3 & 5.8 & 17.6 \\
  HOV-SG~\cite{hovsg}                            &\xmark & Est. &GT &  22.5 & 34.2 & 35.9 & 44.2 & \rd 23.6 & \nd 42.3 & 8.0 & 16.1  \\
  Open3DIS~\cite{nguyen2024open3dis} (SigLip) &\xmark & GT & GT & 25.6 & \nd 38.7 & \fs 49.7 & \fs 64.4 & 22.1 & \fs 42.4 & 4.9 & 9.4 \\ \hdashline \noalign{\vskip 1pt}
  Concept-Graphs~\cite{gu2024conceptgraphs}             &\cmark  & Est. & GT & 16.7 & 33.7 & 27.3 & 39.1 & 15.1 & 35.4 & 4.4  & \fs 26.8 \\
  Open-Fusion~\cite{yamazaki2024open}             &\cmark & Est. & GT & 20.5 & 34.8 & 37.9 & 51.7 & 14.0 & 30.3 & 9.8  & \rd 22.2 \\[0.8pt]  
  \textbf{\ours-mapping (ours)}                          &\cmark & Est. & GT & \nd 27.0 & \fs 39.1 & \nd 45.0 & \nd 59.9 & \fs 25.1 & 38.5 & \rd 11.0 & 18.8   \\ 
  \hdashline \noalign{\vskip 1pt}
  \textbf{\ours-Gaussian-SLAM (ours)}                                         & \cmark & Est. &  \fs 0.6& \fs 27.1 & \rd 38.6 & \rd 44.1 & 58.0 & \nd 25.0 &\rd 39.0 & \fs 12.1 & 18.9   \\ 
  \textbf{\ours-ORB-SLAM2  (ours)}                                     &\cmark & Est. & \nd1.9 & \rd 25.6 &  39.0 & 43.0 & \rd 59.1 & 21.6 & 38.3 & \fs 12.1 & \rd 19.6   \\ 
  \bottomrule
\end{tabular}
\end{table*}
\begin{table*}[ht]
\centering
  \caption{3D semantic segmentation on ScanNetv2 with frequency weighted metrics on 5 (HVS) and all 312 val. scenes (FVS).}
  \label{tab:scannetv2}
\footnotesize
\setlength{\tabcolsep}{4.0pt}  
\begin{tabular}{cl|ccc|ccccccccc}
  \toprule
    &&&\multirow[b]{2}{*}[5pt]{\makecell{Geo-\\metry}}& \multirow[m]{2}{*}{\makecell{Camera pose / \\ATE RMSE [cm]}} & \multicolumn{4}{c}{ScanNet20} & &\multicolumn{4}{c}{ScanNet200} \\ \cline{6-9}  \cline{11-14}
  &Method &Online & & & mIoU & mAcc & f-mIoU & f-mAcc & & mIoU & mAcc & f-mIoU & f-mAcc  \\
  \midrule
  \multirow{10}{*}{
\rotatebox[origin=c]{90}{HVS}} & Open3DIS~\cite{nguyen2024open3dis} (SigLip)   &\xmark &GT&GT& \rd 37.3 & \nd 52.8 & \rd 57.0 & \rd 67.9 && \fs 17.8 & \nd 23.7 &  27.9 & 34.1\\
  
  &OpenScene(Ensemble)~\cite{Peng2023OpenScene}   &\xmark &GT&GT & \fs 44.6 & \fs 61.9 & \fs 57.6 & \fs 71.0 & & 9.4 & 12.6 & 27.8 & 32.0\\
  &HOV-SG~\cite{hovsg}  &\xmark & Est.&GT & 34.4 & \rd 51.1 & 47.3 & 61.8 && 11.2 & 18.7 & 27.7 & 37.6 \\[0.8pt]
  \cdashline{2-14} \noalign{\vskip 1pt}
  &Concept-Graphs~\cite{gu2024conceptgraphs}   &\cmark & Est. &GT& 17.1 & 29.1 & 26.0 & 33.1 & & 6.0 & 11.7 & 21.4 & 27.7 \\
  &Open-Fusion~\cite{yamazaki2024open}    &\cmark &Est. &GT& 30.1 & 39.9  & 54.1  & 68.1  & & 8.6 & 12.8 & \nd 38.4 & \rd 47.9  \\[0.8pt]
  \cdashline{2-14} \noalign{\vskip 1pt}
  %
  &\textbf{\ours-mapping (ours)} &\cmark &Est.& GT&\nd 38.1 & 50.5 & \fs 57.6 & \nd 70.5 && \nd 17.2 & \fs 25.3 & \fs 45.4 & \fs 56.4\\
  &\textbf{\ours-Gaussian-SLAM (ours)}             &\cmark&Est.&\nd{23.7}& 29.3 & 41.1 & 43.0 & 59.5&&  11.8 & 18.8 & 30.1 & 42.6 \\
  &\textbf{\ours-ORB-SLAM2 (ours)} &\cmark &Est.& \fs{21.5}& 31.3 & 45.2 & 45.8 & 61.2 && \rd 13.6 & \rd 22.2 & \rd 38.2 & \nd 51.0 \\
  &{\bf \ours-ORB-SLAM2 w/o loop clos.} &\cmark &{Est.}& \rd{30.2}& {23.6} & {34.5} & {41.4} & {56.9} && {10.3} & {17.3} & {33.2} & {46.0} \\
  \midrule
  \multirow{3}{*}{
\rotatebox[origin=c]{90}{FVS}}  & Open3DIS~\cite{nguyen2024open3dis} (SigLip)  &\xmark &GT&GT& \rd 24.7 & \rd 40.9 & \rd 32.5 & \rd 45.3 & & \rd 9.4 & \rd 17.0 & \rd 22.9 & \rd 32.2 \\[0.8pt]
  &OpenScene(Ensemble)~\cite{Peng2023OpenScene}  &\xmark &GT&GT& \fs 47.0 & \fs 70.3 & \fs 57.7 &\fs 69.8 & & \nd 11.6 & \nd 22.8 & \nd 24.5 & \nd 29.2 \\
  \cdashline{2-14} \noalign{\vskip 1pt}
  & \textbf{\ours-mapping (ours)} &\cmark&Est.&GT& \nd 37.3 & \nd 58.9 & \nd 55.13 & \nd 69.4& & \fs 17.4 & \fs 35.9 & \fs 44.3 & \fs 57.8\\
  \bottomrule
  \end{tabular}
  \vspace{-5pt}
\end{table*}

\boldparagraph{Metrics.} 
Semantic classification is evaluated using \emph{mean Intersection Over Union} (mIoU) and \emph{mean Accuracy} (mAcc) on ScanNet++, while on ImageNet-S we report the standard Top-1 and Top-5 mAcc. While we assess \clipours in 2D to isolate other factors, the full \ours is evaluated in 3D by labeling the vertices of ground-truth meshes and comparing them against ground-truth 3D labels.
For Replica, following OpenNeRF~\cite{engelmann2024opennerf}, we report mIoU and mAcc, categorizing labels into tertiles based on their frequency (\emph{head}, \emph{common}, and \emph{tail}). 
In ScanNetv2, we further present metrics weighted by the label frequency in the ground truth (f-mIoU and f-mAcc).
Additionally, we analyze our computational footprint.
We measure wall-clock time required to optimize Replica scenes, as well as mean and max GPU vRAM and max system RAM usage (in GB).
Each table highlights \colorbox{Green!25}{\textbf{first}}, \colorbox{SpringGreen!45}{second}, and \colorbox{Yellow!30}{third} best results.

\subsection{3D Semantic Segmentation}
\label{sec:sem_seg_exp}
%

\boldparagraph{Replica.} \cref{tab:replica} presents segmentation results for all our \ours configurations alongside relevant baselines. \ours outperforms all baselines in the aggregated mIoU and mAcc (`All' column).
\ours-Gaussian-SLAM and \ours-ORB-SLAM2 surpass both offline and online mapping algorithms.
This is particularly noteworthy since both implementations estimate camera poses and scene geometry, whereas all baselines (indicated in the table) rely either on ground-truth geometry, camera pose, or both.  
Thanks to the strong generalization of our \clipours, all \ours implementations have a significantly better mIoU on \emph{tail} categories, which demonstrates less false-positives.
%
As shown in \cref{fig:replica}, \ours effectively segments and classifies 3D instances, such as chairs and tables, that other baselines often misclassify due to the excessive context information incorporated into CLIP descriptors.
\ours even outperforms the ground truth in some instances.
For example, in "office4" (top left of \cref{fig:replica}), the ground-truth label for the table is missing, and one chair is misclassified as the floor.
This underscores the advantage of open-set pipelines, particularly in situations where previous SLAM algorithms, which rely on known 2D semantics~\cite{li2024sgs, zhai2024nis}, would fail.

\boldparagraph{ScanNetv2.} Results, summarized in ~\cref{tab:scannetv2}, show how \ours-mapping matches HOV-SG, and even Open3DIS in the set ScanNet20.
On the harder set ScanNet200, \ours-mapping has a similar performance to Open3DIS in mIoU, although it is significantly better in terms of f-mIoU and f-mAcc.
OpenScene does achieve the best performance on ScanNet20. 
Nevertheless, its significant drop when using the extended set of classes highlights a weaker generalization capabilities than \ours and other baselines.

\begin{figure}
    \centering
    \includegraphics[width=\linewidth]{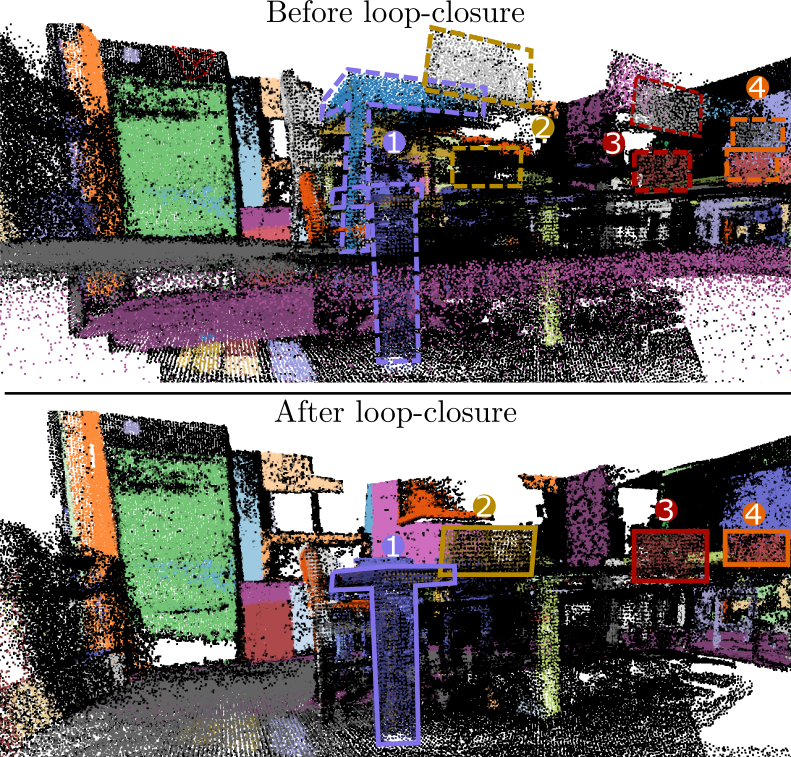}
    \caption{{\textbf{Visualization of OVO-ORB-SLAM2 loop closure on  ``scene0011\_00'' (ScanNet)}. We highlight four instances split due to tracking drift and effectively merged after loop-closure by our semantic fusion.}}
    \label{fig:loop_closure}
\end{figure}

{\boldparagraph{SLAM comparison.}} The difference between \ours's two SLAM versions and \ours-mapping is bigger in ScanNetv2 than in Replica (compare \cref{tab:replica} and \cref{tab:scannetv2}), due to image blur and noisy depths in ScanNetv2.
{Gaussian-SLAM benefits from a more complex strategy for densification and pruning of the 3D point cloud, outperforming our simpler depth unprojection in Replica.
However, while its camera tracking works flawlessly there, it does struggle in ScanNetv2 noisier images, where loop-closure plays a key role.
}
Comparing \ours-ORB-SLAM2 w/o and w/ loop-closure, \cref{tab:scannetv2}, shows the importance of this feature. {Further, \cref{fig:loop_closure} illustrates the loop closure correction over} inconsistent reconstructions with repeated semantic instances, caused by odometric drift.

\begin{table}[tb]
  \caption{Runtime statistics on Replica with 2k frames per scene.}
  \label{tab:time}
  \centering
  \footnotesize
  \setlength{\tabcolsep}{11.7pt}  
  \begin{tabular}{lrrr}
    \toprule
           & vRAM & RAM & Time \\
    Method & Avg / Max & Max & Avg \\
    \midrule
    HOV-SG~\cite{hovsg}                  & \rd 6 / 12 GB &  139 GB &  $\sim$11h\\
    OpenNeRF~\cite{engelmann2024opennerf} & 4 / 22 GB & 44 GB & $\sim$20m\\ \hdashline    
    Open-Fusion~\cite{yamazaki2024open}                &  \fs 3 /\phantom{2} 4 GB & \fs 6 GB & \fs $\sim$3m \\
    CG~\cite{gu2024conceptgraphs}                &  \rd 7 / 11 GB & 16 GB &  $\sim$16m \\
    \ours-mapping (ours)                  & \nd 4 / \phantom{2}8 GB & \rd  12 GB & \nd $\sim$6m \\
    \bottomrule
  \end{tabular}
\end{table}

\boldparagraph{Computational footprint.}
Despite OpenFusion being \(2\times\) faster than \ours, thanks to using SEEM instead of SAM+SigLIP, ~\ours achieves a better balance between speed and performance. It is still \(2.5\times\) faster than Concept-Graphs, $3\times$ faster than OpenNerf and $80\times$ faster than HOV-SG, as shown in \cref{tab:time}. 
In contrast with HOV-SG, that relies on an expensive hierarchical merging of segments, requiring almost \(\times10\) more RAM, ~\ours shows a lower RAM and GPU vRAM usage that enables its use on consumer devices.
{\ours-ORB-SLAM2 takes on average $0.67$ seconds per keyframe on Replica and ScanNetv2, spiking up to $1.4$ seconds for the slowest frame, and up to up to 2.5 and 6.1 seconds after loop closure on ``scene0011\_00'' and ``scene0231\_00'' respectively. This is compatible in our experiments with a conservative keyframe creation policy of 1 keyframe every 10 frames.
}
Therefore, it is compatible with real-time SLAM pipelines, in which the critical camera tracking runs at video rate while the mapping runs at lower frequencies.
%

\subsection{{CLIP Merging}}
\label{sec:clipmerger_eval}
In \cref{tab:clipmerging_imagenet}, we report evaluation on ImageNet-S, including also Alpha-CLIP, and on which both HOV-SG's and Concept-Fusion's merging approaches are equivalent to just computing the global descriptor due to there being only one mask per image.
\cref{tab:clipmerging} presents 2D semantic classification results on unseen scenes from ScanNet++, {using} the expanded label set {of 1.6k}, of our \clipours vs. HOV-SG's and Concept Fusion's (CF) CLIP merging, and the simpler mask crop used by Concept-Graphs. 

\begin{table}[ht!] 
    \caption{ImageNet-S semantic classification accuracy.}
    \label{tab:clipmerging_imagenet}
    \centering
\footnotesize
\setlength{\tabcolsep}{9.2pt}  
\begin{tabular}{lcc}
    \toprule
    Method         &  Top-1 mAcc & Top-5 mAcc \\
    \midrule
    Alpha-Clip (ViT-L/14@336)\cite{sun2024alpha}     & 77.6               & 94.1               \\
    SigLIP-SO400M\cite{zhai2023siglip}     & \nd 82.5               & \nd 95.7               \\
    Mask crop      & \rd 80.3               & \rd 93.4               \\
    \textbf{Our CLIP merging}   & \fs 84.8               & \fs 96.6               \\
    \bottomrule
    \end{tabular}
 \end{table}
\begin{table}[ht]
\vspace{-5pt}
    \caption{2D Open vocabulary semantic classification on ScanNet++.}
    \label{tab:clipmerging}
\centering
\footnotesize
\setlength{\tabcolsep}{7.0pt} 
\begin{tabular}{lcccc}
\toprule
Method & mIoU & mAcc & f-mIoU &  f-mAcc\\\midrule
Concept-Fusion\cite{conceptfusion} & 8.6 & \fs 17.0 &  10.0 &  12.8 \\
Mask crop & 8.5 & \fs 17.0 &  10.0 &  12.8 \\
HOV-SG merging\cite{hovsg} & \nd 9.4 & \rd 15.9 & \rd 12.8 & \rd 15.9\\
\bf Our \clipours  & \fs 9.5 &   14.3 &\nd 36.9 & \nd 49.4 \\
\midrule
{\bf \textit{\clipours variations}} \\
{- fused } &  6.2 &  11.6 & \fs 40.0 & \fs 56.7 \\
{- per-descriptor} &  9.0 & \rd 15.6 &  12.7 & \rd 15.9 \\
\bottomrule
\end{tabular}
\end{table}
\begin{table}[h!]
\centering
    \footnotesize
    \setlength{\tabcolsep}{1.7pt} 
   \caption{{3D Open-Vocabulary Semantic metrics on Replica of OVO-mapping with alternative CLIP merging.}}\label{tab:replica_ablation}
\begin{tabular}{lccccccccc}
\toprule
 &  \multicolumn{2}{c}{All} && \multicolumn{2}{c}{Seen} && \multicolumn{2}{c}{Unseen} \\ \cline{2-3} \cline{5-6} \cline{8-9} 
 &  mIoU & mAcc & &mIoU & mAcc && mIoU & mAcc & \\ 
 \midrule
w/ HOV-SG's fusion & \nd 20.3 & \rd 38.1 && 22.3 & 45.1&& \fs 18.3& \fs 30.9  \\
\textbf{w/ our \clipours} & \fs 27.0 & \fs 39.1 && \fs 36.7 & \fs 54.7 && \nd 16.9 & \nd 22.8\\
\midrule
\bf \textit{w/ \clipours variations} & & &\\
- fused & \rd20.2 & 33.2 && \nd32.2 & \nd52.5 && 0.8 & 1.2\\
- per-descriptor & \rd 20.2 & \nd38.2 && \rd26.6 &\rd48.6 && \rd 13.6 & \nd 27.4\\
\bottomrule
\end{tabular}
\end{table}
Overall, ours outperforms baselines, particularly in frequency-weighted metrics{, although with slightly worse mAcc in ScanNet++.}
Alpha-CLIP performs worse than simpler approaches like our~\clipours. Using a better backbone (SigLIP-SO400M vs ViT-L/14) outperforms a significantly more expensive fine-tuning (our trained two day on 4 V100 vs their on 128 A100 GPUs). 

{Regarding alternatives, the per-descriptor weights predictor achieves a similar performance to HOV-SG, while directly predicting a fused descriptor achieves better frequency-weighted metrics but significantly worse overall ones, which indicates overfitting.}
{This is further validated evaluating OVO-mapping in Replica, \cref{tab:replica_ablation} using HOV-SG's merging approach, and the alternatives to our \clipours. The fused predictor performance collapses in classes not seen during training, while our proposed \clipours has a slightly worse performance than HOV-SG's, while being significantly better on the known classes. 
The per-CLIP weights is unable to match the performance, highlighting the impact of per-dimension weights.}

Finally, we highlight in \cref{fig:gen_queries} how our \clipours preserves their rich semantic encoding, allowing our merged CLIPs to generalize to zero-shot complex language queries.
For instance, our descriptors  distinguish between two trash bins based on a recycling symbol on one of them, despite both being labeled just as \textit{bin} in the ground truth.

\subsection{{Limitations}}\label{sec:limitations}
{Despite OVO state-of-the-art results on 3D indoor semantic segmentation, generalization to outdoor large-scale scenes may face challenges such as different class distributions, illumination and blur, and higher tracking errors.
Our semantic fusion at loop closure effectively corrects odometric drift. However, it sometimes misses instances that should be fused, something that may be fixed by a richer and more accurately localized set of features.
We also observed in \clipours a slight bias towards classes seen at training, which may be solved with larger training sets.}

\section{Conclusions}
In this paper, we present \ours, an open-vocabulary, online 3D mapping method.
Our pipeline extracts 3D segments from 2D masks and tracks them across keyframes.
To assign CLIP descriptors to 3D segments, we introduce a novel strategy: each 2D segment receives a single descriptor computed as a weighted sum of embeddings from the full image, the masked region, and its surrounding bounding box. The weights are predicted by a neural network, which outperforms handcrafted heuristics while retaining strong generalization.
We also develop a mechanism to fuse instances that are affected by odometric drift after the geometric corrections of a loop closure.
\ours outperforms existing baselines in both computational efficiency and segmentation quality across multiple datasets.
By bridging SLAM with open-vocabulary representations, we believe that our work broadens the scope of applications in these two domains.

{
    \small
    \bibliographystyle{IEEEtran}
    \balance
    \bibliography{main.bib}
}
\clearpage
\setcounter{page}{1}
\appendix
\section{Evaluation details}
\subsection{Datasets}
\label{sup:datasets}
\textbf{ScanNet}++ contains \(1752\times1168\) RGB-D images of real indoor scenes with ground-truth 3D meshes and instance and semantic annotations. 
For training, we use the top 100 semantic labels from the more than 1.6K annotated semantic classes, and evaluate on the whole set of 1.6K labels.
Its training set has 230 scenes and its validation set has 50 scenes. 
Each scene has a training camera trajectory and an independent validation one.
\newline
\textbf{ScanNetv2} also images real indoor scenes at RGB resolution of \(1296\!\times\!968\) and depth resolution of \(640\!\times\!480\).
It also has ground-truth 3D meshes with ground-truth instance and semantic annotations.
ScanNetv2 has two sets of annotations, the original set with 20 classes (ScanNet20), and an expanded set with 200 classes (ScanNet200)~\cite{rozenberszki2022language}.
We evaluate on the 5 scenes subset used by HOV-SG~\cite{hovsg} (HVS), and on the whole validation set of 312 scenes (FVS). 
Despite some overlap in physical scenes, ScanNet and ScanNet++ were captured years apart, with different trajectories and sensors, making images and reconstructions significantly different.
Image blur and noisy depths make ScanNet more challenging than ScanNet++.
\newline
\textbf{Replica} is a synthetic dataset generated from high-fidelity real-world data.
Scenes consist of ground-truth 3D meshes with semantic annotations. For all scenes, RGB-D sequences have been rendered at \(1200\!\times\!680\). For Replica we use the common 8 scenes subset (\textit{office-0...4}, \textit{room-0...2}) with NICE-SLAM camera trajectories~\cite{zhu2022nice}.
\subsection{Implementation}
\label{sup:imp}
Our \clipours has a 5-layer transformer encoder with 8 heads and a 4-layer MLP.
It was trained on ScanNet++ train set for 15 epochs, with batch size 512, on 4 V100 GPUs.
As pre-processing, we computed segmentation masks on images, matched these with their ground-truth 2D semantic labels, and pre-computed input and target CLIP embeddings to speed up the training process. 
\newline
Regarding \ours, we use the pixel size of segmented 2D masks as metric of viewpoints quality, and show results selecting the final descriptor between the 10 best keyframes of each 3D segment.
Except when stated otherwise, we relied on SAM2.1-l for 2D instance segmentation, and SigLip ViT-SO400 for CLIP descriptors. 
We query the models with the set of classes of each dataset using the template ``This is a photo of a \{\textit{class}\}''.
For fairness in \ours evaluation, we reproduce previous approaches'~\cite{hovsg, engelmann2024opennerf, takmaz2023openmask3d, Peng2023OpenScene} keyframes selection and querying.
We select as keyframes 1 every 10 frames.
The representation is queried with each dataset's semantic classes, and each 3D segment is matched to the class with higher similarity.
Following HOV-SG, the vertices of our estimated point cloud are matched to the vertices of ground-truth meshes using KD-tree search with 5 neighbors.
Profiling experiments were run on Ubuntu 20, with an i7-11700K CPU, an RTX-3090 GPU, 64 GB of RAM and 150 GB of swap. 
\newline
Due to slight differences in metrics computation, we reproduced HOV-SG and Open3DIS in both Replica and ScanNetv2.
For a fairer comparison with Open-3DIS we implemented it with SigLIP ViT-SO400M rather than its base CLIP ViT-L/14.
We where unable to make OpenNerf converge in ScanNetv2, probably due to the impact of its noisy GT camera poses in NeRFs convergence.
We report OpenNeRF official metrics on Replica.
\begin{figure*}[h!]
  \centering
  \includegraphics[width=\textwidth]{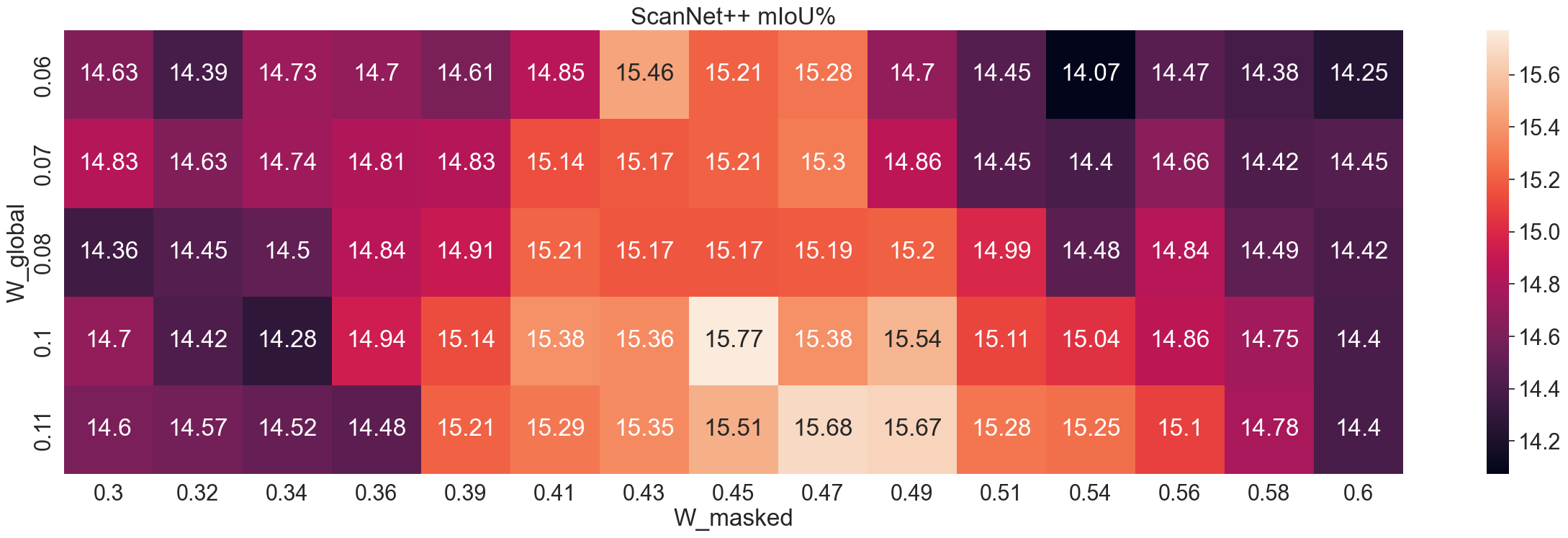}
  \caption{{Grid search for CLIP weights merging on five scenes from ScanNet++~\cite{yeshwanthliu2023scannetpp}.}}
  \label{fig:w_grid_search}
\end{figure*}
\section{System ablations}
In this section we report minor ablations and experiments performed during \ours's development using ScanNet++ training set.
First we report an ablation of different foundation models for 2D instance segmentation, and language-image features extraction.
Then, we ablate the algorithm to merge different CLIP descriptors and validate our proposed \clipours.
We profit from the \clipours to reduce the number of CLIPs descriptors computations and evaluate the impact of the number of views on the selection of the final descriptor of 3D instances.
After that, we present a mask bleeding problem that arises from depth estimation inaccuracies, and how we tackled it.
Finally, we report an overall profiling of the system using different previously ablated components.
\newline
While the segmentation backbones where ablated on a single scene from ScanNet++, we used an extended set of five scenes for CLIP~\cite{radford2021clip} models and similarity computation, to ablate the set of fixed weights, the evaluation of the number of viewpoints, and the mask bleeding.
Then we used a different set of 10 scenes for the overall profiling to avoid overfitting on the previous set.
Regarding \clipours training was done using the 230 scenes from ScanNet++ training set, and validation against baselines was performed on ScanNet++ 50 scenes validation set, and on ADE20K-150.
We measured mean Intersection over Union (mIoU) of the 3D semantic segmentation.
\newline
As starting point, segmentation masks are computed using SAM 2~\cite{kirillov2023sam}; CLIP vectors are computed from masks using SigLIP-384; for each mask three vectors are computed and weighted together as introduced by HOV-SG~\cite{hovsg}; each 3D object gets assigned the CLIP vector from the view that minimizes the L1 distance to its other views. Finally semantic classes are matched to each 3D object using the similarity approach presented by LangSplat~\cite{qin2024langsplat}. 
\subsection{Foundation Models}
\paragraph{SAM} Since its release, Segment Anything Model (SAM)~\cite{kirillov2023sam} has been the state-of-the art for out-of-the box instance segmentation on different fields. Its segment-everything mode extracts multiple masks from a single image, taking an input a grid of point on the image. Nevertheless, this mode has a low throughput mainly due to the post-processing required to filter duplicated and bad segmentation masks.
Although several methods claim up to $\times 100$ speed-ups with respect to SAM, these speed-ups are measured when segmenting a single object on the image, and do not measure the segment-everything mode and its post-processing.
\newline
In this ablation the evaluated models are SAM~\cite{kirillov2023sam}, SAM 2~\cite{ravi2024sam2}, FastSAM~\cite{zhao2023fast}, and EfficientViTSAM~\cite{zhang2024efficientvit}.
The evaluation in ~\cref{tab:ablation_segment} shows how when segmenting everything these methods do not imply an improvement against a SAM implementation with tuned hyper-parameters. 
\begin{table}[h]
\centering
\small
\setlength\tabcolsep{9pt}
\caption{Segmentation backbone ablation.}
\begin{tabular}{lcc}
  \toprule
  & \multicolumn{2}{c}{281bc17764} \\ \cmidrule(lr){2-3} 
  SAM backbone & mIoU$\uparrow$ & Latency [s]$\downarrow$ \\ 
  \midrule
  FastSAM~\cite{zhao2023fast}    & 5.0 & \fs \(0.40\pm0.27\) \\ 
  EfficientViTSAM~\cite{zhang2024efficientvit} & 17.1 & \(4.19\pm0.85\) \\ 
  EfficientViTSAM~\cite{zhang2024efficientvit} - tuned & 15.1 & \nd \(0.68\pm0.05\) \\
  \midrule
  SAM~\cite{kirillov2023sam}         & \nd 19.0& \(5.43 \pm1.83\)  \\
  SAM~\cite{kirillov2023sam} - tuned & \rd 18.1 & \(0.84 \pm0.13\)  \\ 
  SAM 2~\cite{ravi2024sam2} - tuned & \fs 19.1 & \rd \(0.71\pm0.10\) \\ 
  \bottomrule
\end{tabular}
\label{tab:ablation_segment}
\end{table}
\begin{table}[ht]
\centering
\small
\setlength\tabcolsep{3pt}
\caption{CLIP ablation results on 5 scenes from ScanNet++.}
\begin{tabular}{lccc}
\toprule
Architecture & Resolution& mIoU [\%] & Latency [s] \\
\midrule
DFN-ViT-B-16 & \multirow{5}{*}{\(224\times224\)} & 10.92 & \(\fs 0.100\pm0.022\) \\
DFN-ViT-L/14 &  & 11.89 & 
\nd \(0.173\pm0.031\) \\
DFN-ViT-H/14 &  & \rd 13.22 & \(0.286\pm0.054\) \\
OpenCLIP ViT-H/14 &  &  12.71& \(0.283\pm0.053\) \\
SigLIP-SO400M &  & \nd 13.78 & \rd \(0.229\pm0.026\) \\
\midrule
SigLIP-SO400M 384 & \multirow{2}{*}{\(384\times384\)} &  \fs 15.35 & \(0.442\pm0.080\) \\
DFN-ViT-H/14-378 &  &  12.96 & \(0.664\pm0.136\) \\
\bottomrule
\end{tabular}
\label{tab:ablation_clip_arch}
\end{table}
\paragraph{Visual-Language descriptors.}
To compute image-language features we rely on the family of CLIP and its variants.
To select the CLIP architecture we evaluate the difference in performance and latency of different SOTA models to compute CLIP embeddings:
\begin{itemize}
  \item OpenCLIP~\cite{openclip_repo} base ViT-H-14, trained on LAION-2B English~\cite{schuhmann2022laion} at a resolution of \(224\times224\), using CLIP's cosine similarity.
  \item DFN~\cite{data-filtering-networks} ViT-B-16, ViT-L-14, and ViT-H-14 trained on the dataset DFN-5b~\cite{data-filtering-networks} with input images of \(224\times224\), and a ViT-H-14  finetuned at resolution \(384\times384\), using CLIP's cosine similarity.
  \item two SigLIP's Shape-Optimized 400M parameter ViT (ViT SO-400M), trained on WebLI English dataset at \(224\times224\), with one fine-tuned at \(384\times384\), and optimized using SigLIP's cosine similarity.
\end{itemize}
In this ablation each backbone is evaluated using the similarity with which they were trained, without ensembling, and using the template ``\textit{This is a photo of a \{class\}}''.
The results in ~\cref{tab:ablation_clip_arch} show a clear trade-off between segmentation performance, and model latency. SigLIP-384 achieves the best mIoU, while SigLIP at \(224\times224\) has the best balance between mIoU and speed.
Overall, this ablation shows the importance of selecting the proper CLIP backbone, with a difference of almost 5\% between the best and the worst model.
\subsection{CLIP descriptors merging}
\label{sec:supp_CLIP_merging}
\begin{figure*}[ht]
  \centering
    \includegraphics[width=0.9\textwidth]{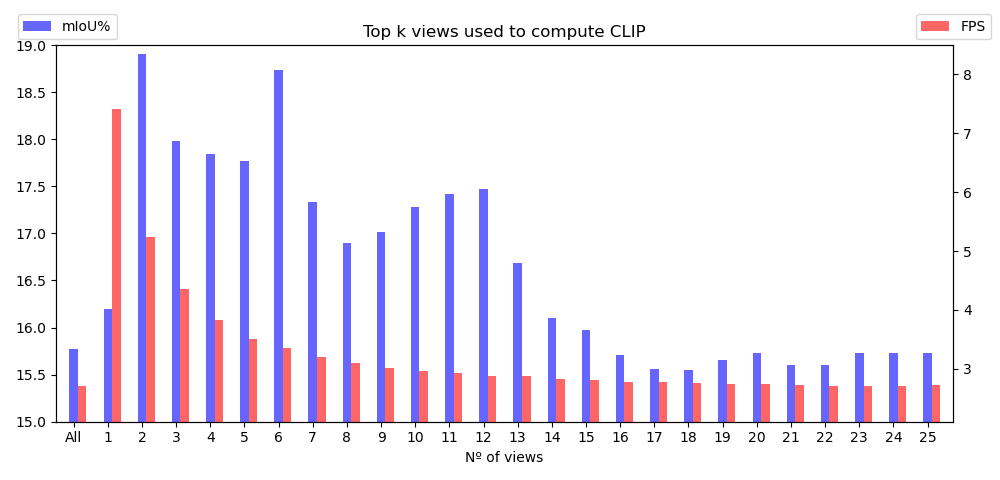}
  \caption{\textbf{Evaluation using only top views to compute CLIP on 5 scenes from ScanNet++~\cite{yeshwanthliu2023scannetpp}.} While using more than one view has substantial impact on the runtime, it also improves segmentation accuracy. However, too many views also degrade the segmentation accuracy. }
  \label{fig:topk}
\end{figure*}
\paragraph{Similarity computation.}
Initially, CLIP~\cite{radford2021clip} presented the cosine similarity, \(\cos\left(\phi_\text{qry},\phi_\text{img}\right)\) to compute the distance between the text, $\phi_\text{qry}$, and image, $\phi_\text{img}$, embeddings. SigLIP~\cite{zhai2023siglip} adapted it to its loss function, as \(\mathrm{Sigmoid}\left(\cos\left(\phi_\text{qry},\phi_\text{img}\right)\times \tau + b \right)\), including a Sigmoid operation, and the learned inverse temperature, \(t = \frac{1}{\tau}\), and bias \(b\) parameters. To classify, both approaches assigned to an image the class of the query that generated the highest similarity.
Also based on CLIP's experiments, to compute the cosine similarity HOV-SG~\cite{hovsg} computed query embeddings as \(\phi_\text{qry} = \frac{\phi_\text{cl}+\phi_\text{temp}}{2}\), where $\phi_{cl}$ is the text embedding computed from the class name, and $\phi_\text{temp}$ is the text embedding computed from the phrase resulting of inserting the class into the template ``\textit{There is \{class\} in the scene}''.
In contrast, LERF~\cite{lerf2023} proposed to compute the cosine similarity between the image and text embeddings as 
\begin{equation}
  \min_i \frac{\exp\big(\cos\left(\phi_\text{qry},\phi_\text{img}\right)\big)}{\exp\big(\cos\left(\phi_\text{qry},\phi_\text{img}\right)\big) + \exp\big(\cos\left(\phi_\text{can}^{i},\phi_\text{img}\right)\big)} \, ,
\end{equation}
where $\phi_\text{can}^{i}$ is the text embedding of one of the predefined canonical queries \textit{object, things, stuff, texture}.

\noindent
Using the SigLIP ViT-SO400M model to compute CLIP vectors, we compare between:
\begin{itemize}
  \item computing query embeddings, $\phi_\text{qry}$, only with the template  ``\textit{This is a photo of a \{class\}}'' or as an ensemble averaging the template embedding with the class embedding;
  \item and computing SigLIP's cosine similarity or LERF's cosine similarity.
\end{itemize}
Results in ~\cref{tab:ablation_sim} show how the basic configuration of using SigLIP similarity without ensemble achieves the best performance.
\textbf{From here on, all experiments will proceed using basic cosine similarity without ensemble.}
\begin{table}[]
\centering
  \caption{Similarity computation ablation on 5 scenes from ScanNet++ measuring semantic 3D mIoU.}
  \begin{tabular}{lcc}
    \toprule
                & Cosine similarity & LERF's similarity \\ \hline
  w ensemble    & 14.75\%           & 14.75\%         \\
  w\textbackslash o ensemble & \textbf{15.35\%}           & \bf 14.98\% \\ \hline       
  \end{tabular}
  \label{tab:ablation_sim}
\end{table}
\begin{table*}[]
    \centering
    \small
    \caption{Our \clipours vs. baselines on: ScanNet++ (S++) using 1.6k queries (metrics on observed 495 labels, w. and w/o. the top 100 used at training), and ADE20k with 150 labels. Color indicates \colorbox{Green!25}{First}, \colorbox{SpringGreen!45}{second}, and \colorbox{Yellow!30}{third} best.}
\begin{tabular}{lccccccccccccc}
\toprule
 &\multicolumn{4}{c}{S++ w. top 100} &\multicolumn{4}{c}{S++ w/o. top100}&\multicolumn{4}{c}{ADE20k--150}\\ 
Method & mIoU & mAcc & f-mIoU &  \multicolumn{1}{c|}{f-mAcc} & mIoU & mAcc & f-mIoU &  \multicolumn{1}{c|}{f-mAcc} & mIoU& mAcc  & f-mIoU & f-mAcc\\
\midrule
HOV-SG & \nd 9.4 & \nd 15.9 & \rd 12.8 &  \multicolumn{1}{c|}{\rd 15.9}  & \fs 8.3 & \fs 15.1 & \nd 8.4 &  \multicolumn{1}{c|}{\rd 13.6} & \rd 21.9 & \nd 53.7& \rd 22.3  & \rd 34.9\\
Fixed-weights & \nd 9.4 & \nd 15.9 & \nd 13.1 &  \multicolumn{1}{c|}{\nd 16.3}  & \fs 8.3 & \fs 15.1 & \nd 8.4 &  \multicolumn{1}{c|}{\nd 13.8} & \nd 22.4 &\fs 53.9 & \nd 23.1 &\nd 35.5 \\
\bf CLIP-merger & \fs 10.7 & \fs16.9 &\fs 36.1 &  \multicolumn{1}{c|}{\fs 45.3} & \nd 7.3 & \nd 12.8 &\fs 9.9 &  \multicolumn{1}{c|}{\fs15.0} & \fs 23.4 & \rd 49.3 & \fs 28.7 & \fs 41.2 \\
\bottomrule
\end{tabular}
    \label{tab:clipmerger2}
\end{table*}
\begin{table*}[]
\centering
\small
\setlength{\tabcolsep}{6.5pt} 
  \caption{\textbf{Average runtimes and 3D semantic performance on ScanNet++.} We measure the segmentation (Seg); segments matching and tracking (M\&T); segments pre processing (PP); CLIPs computation (CLIP); and total seconds per key frame (\(s/KF\)). 
  }
\begin{tabular}{clc|cccc|c|ccccc}
  \toprule
  CLIP & SAM & \# best views & Seg. [s] & M\&T [s] & PP [s] & CLIP [s] & \(s/KF\) & mIoU & mAcc & f-mIoU & f-mAcc \\
  \hline
  \multirow{2}{*}{ViT-H/14} & 1-H & \multirow{2}{*}{10}  & 1.516  & 0.269  & 0.085  & \nd 0.175  & 2.112  & 13.3 & 22.4 &20.2 & 31.7 \\\cline{2-2}
  &  2.1-L & & \rd 0.338  & \nd 0.252  & \rd 0.066  & \fs 0.135  & \nd 0.865  & \rd 14.1 & 24.9 &  27.3 & 37.7 \\\midrule
  \multirow{3}{*}{SigLIP} &  2.1-t &  \multirow{2}{*}{10} & \fs 0.245 & \fs 0.247  & \fs 0.057  & \rd 0.204  & \textbf{0.820}  & 11.8 & \rd 25.7 & \rd 34.2 & 
  \nd 46.6  \\ \cline{2-2}
  &  \multirow{2}{*}{2.1-L} & & 0.339  & \rd 0.253  & \nd 0.065  & 0.233  & \rd 0.957  & \nd 14.2 & \nd 27.0 & \nd 34.3 & \rd 45.6 \\\cline{3-3}
  &    & all & \nd 0.337  & 0.261  & 0.110  & 0.367  & 1.167  & \fs 15.8 & \fs 29.6 & \fs 36.3 & \fs 48.6 \\ 
  \hline
  \end{tabular}
  \label{tab:ablation_backbones}
\end{table*}
To focus CLIP descriptors to elements in an image, we follow HOV-SG's~\cite{hovsg} approach.
For each mask segmented by SAM, HOV-SG proposed to compute CLIP embeddings combining the information of the complete image, the masked image without background, and a bounding box of the mask including background.
For each segmentation mask $i$, its corresponding CLIP vector $F_i$ is computed as
\begin{equation}
F_{i} = F_\text{global} \times w_\text{global} + F_{\text{local}_i} \times (1 - w_\text{global}),
\end{equation}
with 
\begin{equation}
F_{\text{local}_i} = F_{\text{masked}_i} \times w_\text{masked} + F_{\text{bbox}_i} \times (1-w_\text{masked}),
\end{equation}
combinig the CLIP vector of the whole image, $F_\text{global}$, the CLIP vector of only the segmentation mask without background, $F_{\text{masked}_i}$, and the one of the bounding box of the segmentation mask including background, $F_{\text{bbox}_i}$.

HOV-SG~\cite{hovsg} used
\begin{equation}
  w_\text{global}= \mathrm{Softmax}(\cos(F_\text{global},F_{i})),
\end{equation}
and $w_\text{masked} = 0.4418$. 
Nevertheless, the use of the $\mathrm{Softmax}$ introduced a dependency between the different embeddings extracted on the same frame. To avoid computing all CLIP embeddings on every frame, we remove the $\mathrm{Softmax}$ and perform a grid search of \(w_\text{masked}\) and \(w_\text{global}\).
The best performance is achieved for \(w_\text{global} = 0.45 \text{ and } w_\text{masked} = 0.0975\) as shown in ~\cref{fig:w_grid_search}.
\paragraph{\clipours}
Rather than relying on 3 fixed-weights that ideally should be tunned for each scene, we developed the \clipours to estimate the corresponding weight for each image.
After training on ScanNet++ train set with the top 100 semantic labels, we evaluate its performance on the ScanNet++ validation set using the total set of 1.6k queries, both including (w.top 100) and excluding (w/o. top 100) classes seen during training. For a stronger distribution switch, we also evaluate on ADE20k-150.
\newline
Comparing its performance against HOV-SG's approach and our variation of HOV-SG's using three fixed weights, the \clipours outperforms the baselines using all the labels, \cref{tab:clipmerger2}. 
Excluding from the metrics the 100 labels seen during training, we can observe how the \clipours performance drops with respect to the baselines.
Despite the slight bias toward classes at training, it still outperform on freq. weighted metrics of classes that weren't seen during training, and on novel data on the ADE20k-150 dataset.
\newline
Although, \ours-mapping evaluation in Replica and ScanNetv2, additional segmentation metrics on classes outside the training set (\cref{tab:miou_macc} showcase how the bias does not have an impact on our \clipours's generalization.
Our method accurately detects in 3D several unseen classes across Replica and ScanNetv2, including \textit{guitar}, \textit{coffee maker}, \textit{blackboard}, and \textit{scale}. The mIoU for these examples exceeds 60\%. 
\textbf{From here on, all experiments will proceed using the \clipours.}
\begin{table}[t]
\centering
    \small
    \setlength{\tabcolsep}{3pt} 
    \caption{\textbf{\clipours generalization.} 3D metrics on ScanNetv2 of some classes not seen during training.}\label{tab:miou_macc}  
\begin{tabular}{lcccccc}
\toprule
 & \multirow{2}{*}{scale} & toaster  & \multirow{2}{*}{blackboard} & coffee  & \multirow{2}{*}{guitar} & projector  \\ 
 &  & oven &  & maker &  &  screen \\ \hline
mIoU\% & 75.1 & 78.53 & 61.4 & 67.0 & 62.68 & 64.1 \\
mAcc\% & 81.2 & 94.07 & 76.1 & 86.7 & 86.79 & 86.8 \\
\bottomrule
\end{tabular}
\end{table}
\subsection{Additional heuristics}
\begin{figure}
    \centering
    \includegraphics[width=\linewidth]{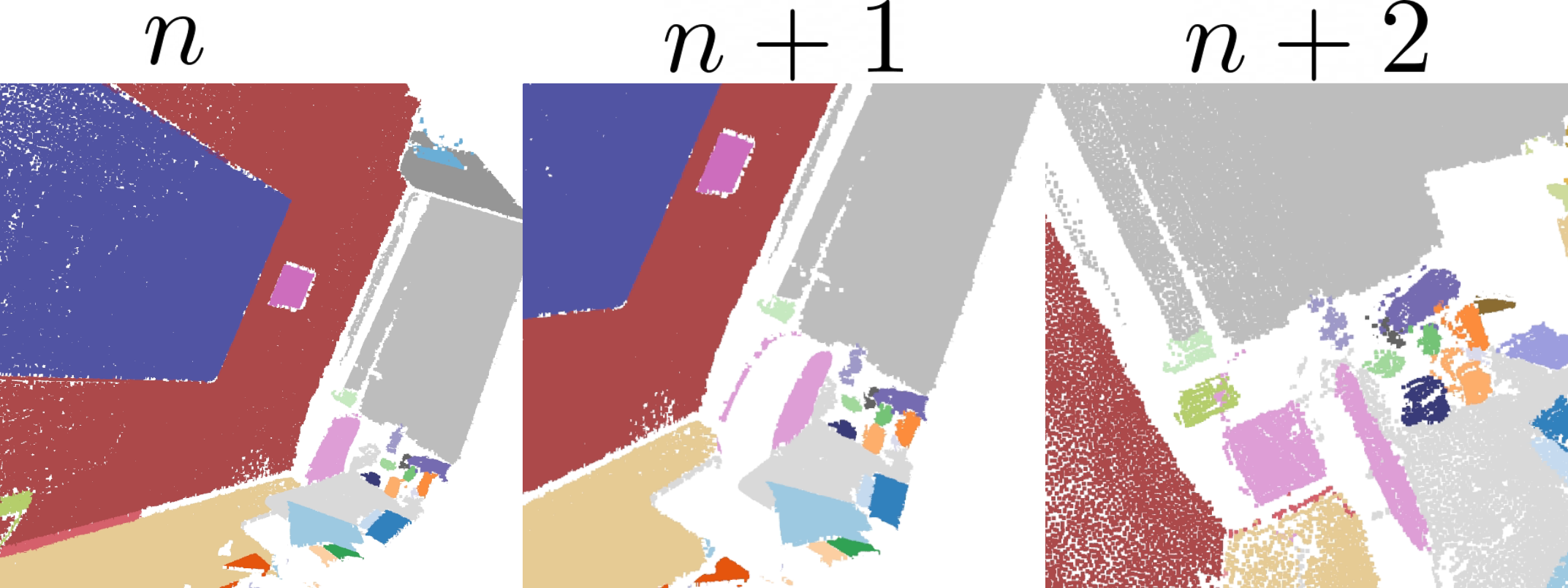}
    \caption{\textbf{Mask bleeding} and propagation produced by masks inaccuracy. The edges of the chair (pink) bleed to the background at $k_n$, and therefore the segment label is wrongly propagated to the it in the following keyframes.}
    \label{fig:mask_bleeding}
\end{figure}

\paragraph{Nº of best views.}
To reduce the expensive CLIP computation for each frame, we evaluate the impact of using only the best views where each 3D segment has been seen to compute its CLIP descriptor.
We evaluate from using only the best image to using all the images where the object has been seen.
The quality of an image is based on the area of the object's 2D segmentation in it.
\newline
For a sequence of 51 keyframes, we evaluate for \(k\in\{1,\ldots,51\}\), being \textit{all} using all the views to compute objects 3D vectors. 
The results show, see ~\cref{fig:topk}, that neither using only the best nor using all the views are robust enough to noise. 
For the set of 5 scenes on this experiment, the best values of \(k\) are between 2 and 7, achieving an mIoU around 18\%, almost 3 points better than using all observations, although, the perfect value of will probably be scene and object dependent.
We decide to set use 10 views as a balance to avoid useless computation of CLIP vectors and being resistant to noisy images.
\paragraph{Masks bleeding.}
Observing OVO-SLAM matching results, we noticed some problems related with SAM's masks. When some 3D points are projected on the edges of a 2D mask to which they do not belong, they are wrongly clustered into it and matched to a 3D instance. Then, when these  are seen again they will propagate the wrongly assigned ID.
This phenomenon can be observed in particular on the edges of objects, where the depth and masks are less accurate, and masks propagate the ID of the object to the background, as seen in~\cref{fig:mask_bleeding}.
To compensate it we developed two approaches:
\begin{itemize}
  \item First, we add a filter to only keep matches of 3D points that are assigned to the same object in two consecutive frames;
  \item Second, we apply a low-pass filter to the depth map to mask the edges of the objects and avoid matching points around them.
\end{itemize}
Results on ~\cref{tab:ablation_depth_kf} show how while using the depth filter does improve the average mIoU, the limitation to match in consecutive frames does not. As a consequence we keep only the depth filter although it does not completely solve the problem.
\begin{table}[ht]
  \centering
  \setlength\tabcolsep{28pt}
  \caption{{Mask bleeding solutions' ablation on 5 scenes from ScanNet++~\cite{yeshwanthliu2023scannetpp}.}}
  \begin{tabular}{lc}
    \toprule
    Config & mIoU$\uparrow$ \\ 
    \midrule
    Base                    & \rd 15.80\% \\ 
    w depth filter          & \fs 16.16\% \\
    w consecutive KF filter & 15.07\% \\
    w both                  & \nd 15.82\%   \\
    \bottomrule
  \end{tabular}
  \label{tab:ablation_depth_kf}
\end{table}
\paragraph{Overall profiling.} 
Finally, we quantify the latency-quality trade-off in our architecture evaluating selected foundation models and number of views against less powerful alternatives.
This evaluation is performed on a different set of 10 scenes from ScanNet++ to avoid over-fitting to the previous 5 scenes.
For 2D segmentation we evaluate SAM~\cite{kirillov2023sam} with ViT-H/14 encoder (1-H), and SAM 2.1~\cite{ravi2024sam2} with Hiera large (2.1-L) and Hiera tiny (2.1-t) image encoders. 
For CLIP extraction, we evaluate DFN ViT-H/14-378~\cite{data-filtering-networks} and SigLIP-SO400~\cite{zhai2023siglip} both with input images of 384 pixels.
The results in \cref{tab:ablation_backbones} show that in this set of scenes the best 3D segmentation is achieved with the largest models using all points of view.
Nevertheless, the best trade-off can be achieved reducing the number of views and the CLIP model.

\end{document}